\documentclass{IEEEtran}
\usepackage{times}

\usepackage[numbers]{natbib}
\usepackage{multicol}
\usepackage[bookmarks=true]{hyperref}

\usepackage[T1]{fontenc}
\usepackage[latin9]{inputenc}
\usepackage{amstext}
\usepackage{amssymb}
\usepackage{graphicx}
\usepackage{esint}

\makeatletter

\newcommand{\noun}[1]{\textsc{#1}}
\providecommand{\tabularnewline}{\\}

\let\vec\b

\makeatother

\usepackage[english]{babel}
\begin{document}

\title{Learning Relevant Features for Manipulation Skills using Meta-Level
Priors}

\author{Oliver Kroemer and Gaurav S. Sukhatme \\ University of Southern California}
\maketitle
\begin{abstract}
Robots can generalize manipulation skills between different scenarios
by adapting to the features of the objects being manipulated. Selecting
the set of relevant features for generalizing skills has usually
been performed manually by a human. Alternatively, a robot can learn
to select relevant features autonomously. However, feature selection usually requires
a large amount of training data, which would require many demonstrations.
In order to learn the relevant features more efficiently, we propose
using a meta-level prior to transfer the relevance of features from
previously learned skills. The experiments show that the meta-level prior
more than doubles the average precision and recall of the feature selection when
compared to a standard uniform prior. The proposed approach was used
to learn a variety of manipulation skills, including pushing, cutting,
and pouring. 
\end{abstract}

\section{Introduction\label{sec:Introduction}}

\begin{figure*}
\begin{centering}
\includegraphics[width=1\textwidth]{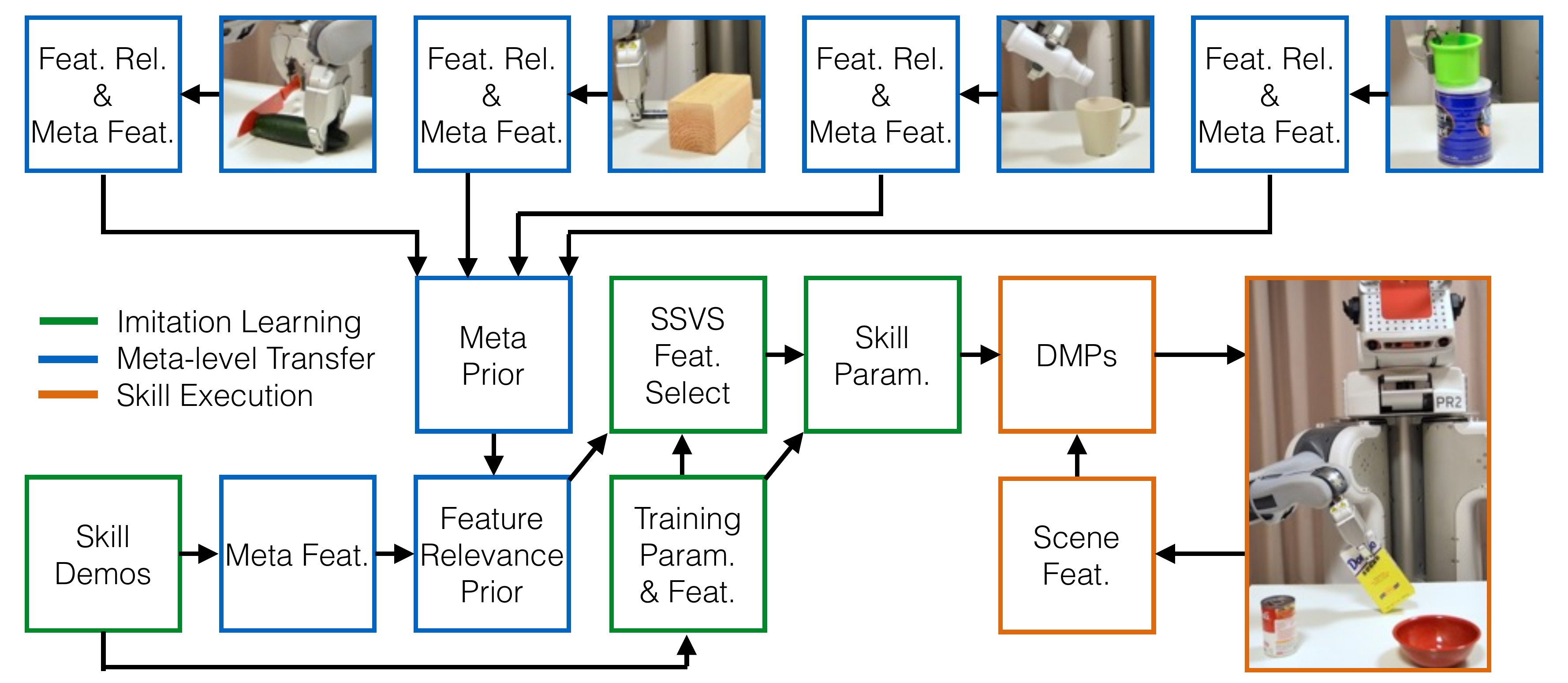}
\par\end{centering}
\protect\caption{ \label{fig:overview} The figure shows an overview of the proposed approach. The green blocks correspond to learning the skill parameters from demonstrations. The orange blocks correspond to executing the skill in novel situations using the learned parameters. The blue blocks correspond to learning the meta-level prior, which can be used to compute a feature relevance prior for the new skill based on the meta features extracted from the skill demonstrations.  }
\end{figure*}

In the future, we want robots to perform a wide range of manipulation
tasks in everyday environments. These robots will need to be capable
of learning new skills from demonstrations and generalizing these
skills between different objects and scenarios.
Variations in objects and their state can be represented by object features.
Object features could, for example, include the 3D position and size of a container's 
opening. Using these features, the robot can learn to generalize a pouring skill between
cups and bowls at different positions in the robot's workspace. Similarly,
a robot could adapt a cutting skill using the position and length features of a knife's edge.
A robot can thus represent task scenarios using geometric features of
affordance-bearing parts.

Many of the extracted object features will however not be relevant for generalizing the manipulation skill.
One of the key challenges for learning manipulation skills in unstructured environments
is therefore to determine which object features are relevant for adapting and
generalizing the skills. One approach would be to have the human demonstrator
manually label the relevant features. However, this approach would
require a considerable amount of additional effort from the human
demonstrator and limit the robot's autonomy. Instead of relying on
prior human knowledge, the robot should utilize information about relevant
features from skills that it has previously learned. Using
this information, the robot can autonomously create a prior regarding
which features are relevant for generalizing novel manipulation skill. 

Predicting the relevance of an object feature is not a trivial task. 
As there is no one-to-one mapping between features of different
tasks, the robot will need to generalize relevance between different
features. This approach may seem impossible if we consider a feature
as simply being a number. However, for manipulation tasks, the features
are generally grounded in the geometry and properties of the objects
being manipulated. Hence, we can define meta features that capture
characteristics of the features and how they relate to the skill being
learned. Using these meta features, the robot can learn a meta prior for computing
a prior over the relevance of features in novel manipulation tasks
\cite{Lee_ICML_2007}. This prior can then be incorporated in the feature
selection process in order to determine the relevant features from
demonstrations more efficiently. 

In this paper, we investigate learning meta-level priors for generalizing
the relevance of features between different manipulation skills. An overview
of the proposed approach is shown in Fig. \ref{fig:overview}.  In order to 
learn a new versatile manipulation skill, the robot is first provided with multiple 
demonstrations of the task using kinaesthetic teaching. The robot is also
provided with 3D point cloud models of the manipulated objects. The robot 
uses these demonstrations and models to extract a set of object features 
and their corresponding skill parameters for training the skill. The ultimate
 goal of the proposed framework is to select a sparse set of relevant object 
features and to learn the corresponding parameters for adapting the skill execution accordingly. 
The part-based feature generation process is explained in Section~\ref{sec:Generating-Object-Features}, and the 
parameterized skill representation is described in Section~\ref{sub:Motor-Primitive-Skill}.
 
For each object feature, the robot also extracts a set of meta features, which describe
 the relationship between the feature and the skill being learned. These meta 
features are used to compute a prior over the relevance of the feature. 
For example, as each object feature is associated with an object part, one meta feature
may describe the distance between the robot's hand and the feature's part at 
the start of the demonstration. Another meta feature may represent 
this distance at the end of the demonstration. Using these two  meta feature, 
the robot may learn that a feature is more likely to be relevant if the robot's 
hand moves closer to the feature's part during the task.  The mapping from 
meta features to the prior over the feature's relevance is based on a meta
prior. The meta-level prior is learned using the meta features and corresponding 
feature relevances from previous skills, as illustrated in the top row of Fig. \ref{fig:overview}. 
The meta features and meta-level prior are explained in Sections \ref{sub:Object-Features-and-meta-features} and \ref{sub:Probablisic-Model-for} respectively. 

Given the training features and skill parameters, as well as the feature relevance prior,
the next step is to autonomously select a sparse set of features for generalizing the
skill between scenarios. The feature selection is performed using stochastic search 
variable selection (SSVS) \cite{George_ASA_1993}. We extend this method to include
the meta-level prior into the feature selection process. The set of relevant features are inferred
using a Gibbs sampling approach. The feature selection process is explained in 
Sections \ref{sub:Probablisic-Model-for} and \ref{sub:Inferring-Relevant-Object}. 
The robot subsequently learns a set of skill parameters for the selected features. 

When presented with a new scene, the robot first computes the set of object features
for the current scenario. It then adapts the skill execution based on the learned parameters
in order to perform the manipulation task. The proposed
framework was evaluated on placing, pushing, tilting, cutting, pouring, and wiping tasks 
using a PR2 robot. The experiments are described and discussed in Section~\ref{sec:Experiments}. 
Using the meta-level prior, the robot was able to more than double  the average
precision and  recall of the feature selection.

\section{Related Work\label{sec:Related-Work}}
Recent work in both imitation learning and policy search reinforcement
learning have allowed robots to learn and execute complicated motor skills.
Example skills include playing ball games \cite{Kober_ICRA_2010,Amato_ICRA_2014,Peters_IROS_2006,Kupcsik_AI_15},
opening doors \cite{Kalakrishnan_IROS_2011,Konidaris_AAAI_2011},
scrubbing surfaces \cite{Do_ICRA_2014,Chebotar_IROS_2014}, and manipulating
objects \cite{Pastor_ICRA_2011,Levine_ICRA_2015,Calinon_IROS_2013}. These learned skills
adapt to different features of the task, e.g., the position of a ball
or the goal location. However, the features used to adapt the skill
executions are usually predefined and do not include irrelevant features.
These methods therefore focus on determining how to adapt to the features,
and not learning which features to adapt to. 

Some frameworks learn relevant features in order to select actions
more accurately in otherwise ambiguous situations \cite{Piater_IJRR_2011,Levine_arxiv_2015}.
These approaches often perform an implicit pose estimation of an object
or part, which the robot can then use to adapt its actions. However,
the features are usually learned for specific sets of objects and not to generalize between objects with different shapes. 

Motor skills can also be adapted to different situations by selecting
suitable task frames. These task frames are often defined relative
to objects for manipulation tasks and, hence, there are usually multiple
potential task frames to choose from. The robot can learn to select
a task frame from multiple demonstrations based on the the variance
of the trajectories in each potential task frame \cite{Pais_HRI_2013,Niekum_AAAI_2013}.
Task frames can be generalized between different sets of objects by
associating the coordinate frame to object parts with similar shapes
\cite{Stuckler_Humanoids_2014,Bohg_TRO_2014,Kroemer_ICRA_2012}. In these cases,
learning the task frame is approached as a part detection and pose
estimation problem. Although task frames are fundamental to generalizing
manipulations, many skill adaptations rely on additional information
regarding the size and shape of the objects being manipulated \cite{Tenorth_IROS_2013}.

Another important challenge for skill learning is determining which
objects are involved in a manipulation \cite{Kemp_RAM_2007}. The set
of relevant objects can often be extracted from demonstrations using
visual cues, such as motionese \cite{Lee_IROS_09}. The object selection
problem is however distinct from the feature selection problem addressed
in this paper, and not all of the features associated with an object
will be relevant for generalizing a skill. 

Several works have investigated transfer and multi-task learning in
the field of robotics \cite{Stolle_IROS_2007,Metzen_GJAI_2013,Deisenroth_ICRA_2014}.
These approaches often focus on transferring trajectories or controllers
between different tasks. The different tasks are also often quite
similar and share the same feature space, e.g., reaching for different
locations in the task frame may be considered as different tasks.
In our work, the tasks have distinct sets of features and the robot
is learning a meta-level prior for transferring the relevance of features
between skills. Meta features have previously been used to transfer
knowledge about the relevance of features between tasks in applications
such as predicting movie ratings, text classification, and object recognition \cite{Lee_ICML_2007,Taskar_ICML_2003,Krupka_JMLR_2008}.

Our feature generation process is based on decomposing objects into
parts. The segmentation is guided by the demonstrations of the skill.
Several other works have also used actions in order to improve segmentation
performance \cite{Hermans_IROS_2012,Hoof_TRO_2014,Hausman_ICRAWS_2012}.
However, most of these methods focus on segmenting scenes into objects
rather than parts. Decomposing objects into parts has largely been
based on geometric features, such as convexity \cite{Bogdan_ICCVWS_09,Stein_ICRA_2014,Goldfeder_ICRA_2007}.
Lakani et al. recently proposed a method for segmenting objects into
parts based on where it could be grasped by the robot \cite{Lakani_CRV_2015}.
Their approach uses an initial over-segmentation of an object model
into supervoxels and merges these supervoxels into parts.

\section{Generating Features and Meta Features\label{sec:Generating-Object-Features}}
\begin{figure}
\begin{centering}
\includegraphics[width=1\columnwidth]{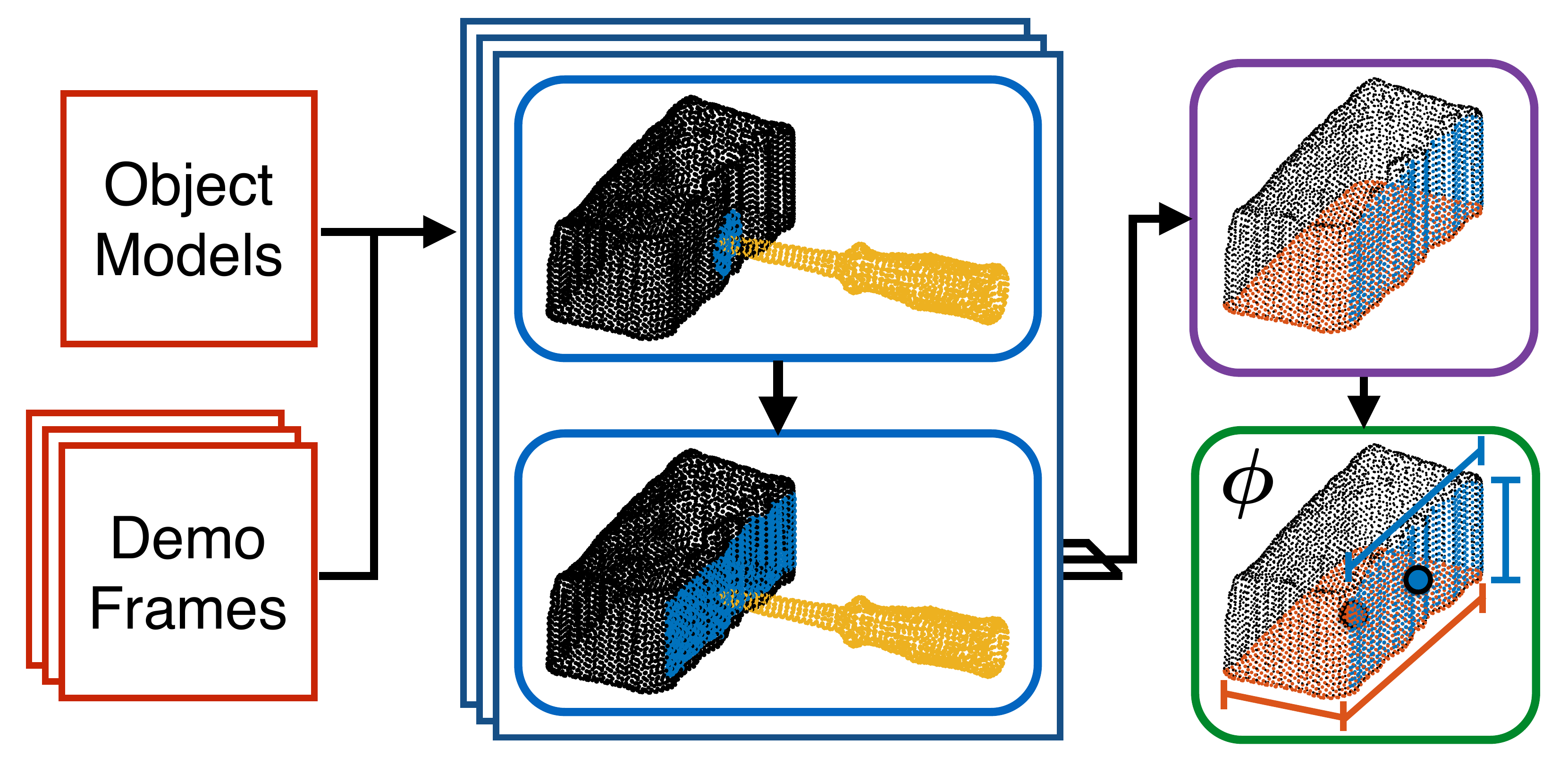}
\par\end{centering}

\protect\caption{ \label{fig:FeatGen} The figure illustrates the key steps of the feature generation process. (Red) The inputs to the process are a set of 3D point cloud models of the objects being manipulated, and a set of scene frames from a demonstration of the task. (Blue) For each frame and pair of objects, the robot estimates a contact region. Using this region as an initialization, the robot computes an estimate of the affordance-bearing part using GrabCut segmentation. (Purple) The individual estimates of the object parts are clustered in order to remove redundant and noisy part estimates. (Green) The robot uses a predefined set of rules to generate three position and three length features for each part. The x-y-z positions and lengths are defined in the task frame. }
\end{figure}

In order to learn a skill that generalizes between scenarios, the
robot first requires a suitable method for representing the scenarios
that it encounters. In particular, the robot must generate a set of
features describing the objects and their current configuration.
Generalizing between objects is often easier when the representation 
explicitly models the manipulation-relevant parts of the objects \cite{Sung_ISRR_2015,Kroemer_ICRA_2012}.
Our feature generation approach therefore begins by identifying parts
of objects that may be relevant to the task. The part segmentation is 
based on the skill demonstrations in order to extract task-relevant parts. The part detection
method is described in Section \ref{sub:Detecting-Affordance-bearing-Parts}.
For each of the resulting parts, the robot generates a set of features describing 
the position and size of the part. An overview of the feature generation process
is shown in Fig. \ref{fig:FeatGen}.

Each of the resulting features is associated with a set of meta features, which are used
to predict the relevance of the feature for the new task. 
The object features and their meta features are described in Section
\ref{sub:Object-Features-and-meta-features}.

\subsection{Detecting Affordance-bearing Parts from Demonstrations\label{sub:Detecting-Affordance-bearing-Parts}}

Affordance-bearing parts are important because they indicate regions
of an object that can be used for manipulating or interacting with
other objects \cite{Gibson_1979}, e.g., the edge of a knife for cutting or the handle
of a cup for grasping. Most of these interactions involve direct physical
contact or close proximity between the objects. The affordances of
a part also depend on its shape, with similarly-shaped regions
of objects having similar affordances. 

Given these two insights, we propose extracting affordance-bearing
object parts using the GrabCut segmentation algorithm \cite{Rother_SIGGRAPG_2004}.
GrabCut is commonly used to segment the foreground from the background
in photo editing. The segmentation iterates between modeling the pixels'
features in each segment and segmenting the pixels into two groups using
a min-cut approach. The distribution over the pixel assignments  is modeled 
as a Markov random field. In the photo-editing domain, the segmentation
is initialized by a human user providing a coarse estimate of points
that belong to the foreground.

In our approach, the robot uses GrabCut to segment the task-relevant
part from the rest of the object. Given a 3D point cloud model, the
robot computes the position, the normal, the curvature, and spectral
features for each point based on its local neighbourhood \cite{Munoz_ICRA_2009,Boularias_IROS_2011}.
The part's affordances may depend on other properties, 
e.g., material and texture, but these are beyond the scope of this paper.
The robot also creates a k-nearest neighbour graph connecting the
points. The unary potentials are modeled using Gaussian mixture models
over the local geometric features. For the pairwise potentials, we
employ the commonly used Potts model \cite{Potts_1952}. GrabCut 
allows for observation-dependent pairwise potentials, but the
model then does not represent a Markov random field prior \cite{Prince_2012}.

An object part is relevant to the task if it is used to interact with
other objects. The segmentation was therefore initialized by selecting
points that were likely to be interacting with other objects during
the demonstrations. As many interactions are based on direct physical
contact, or close proximity, we considered a point to be interacting
if it was within a threshold distance of the points of another object
with opposing normals. This spatial proximity assumption has been
shown to be applicable to a wide variety of manipulation tasks \cite{Aksoy_IJRR_2011}. 

Given a set of object models and their trajectories during a demonstration,
we begin by computing the GrabCut segmentation for each pair of objects
for each frame in which they are in close proximity. This process generates
multiple redundant estimates of the object's parts. In order to merge
the part estimates across the frames, we cluster the part
estimates using spectral clustering \cite{Chung_AMS_1997}. In our experiments,
we computed the similarity between the parts using a Bhattacharayya
kernel over the parts' positions and normals in the object frame \cite{Jebara_COLT_2003}.
The clustering was performed multiple times with different numbers
of clusters, and we selected the clustering that maximized the intra-cluster
kernel values minus the inter-cluster kernel values. The final part
included all of the points that were present in the majority of the
cluster's estimates.

\subsection{Object Features and Meta Features\label{sub:Object-Features-and-meta-features}}

\begin{figure}
\begin{centering}
\includegraphics[width=1\columnwidth]{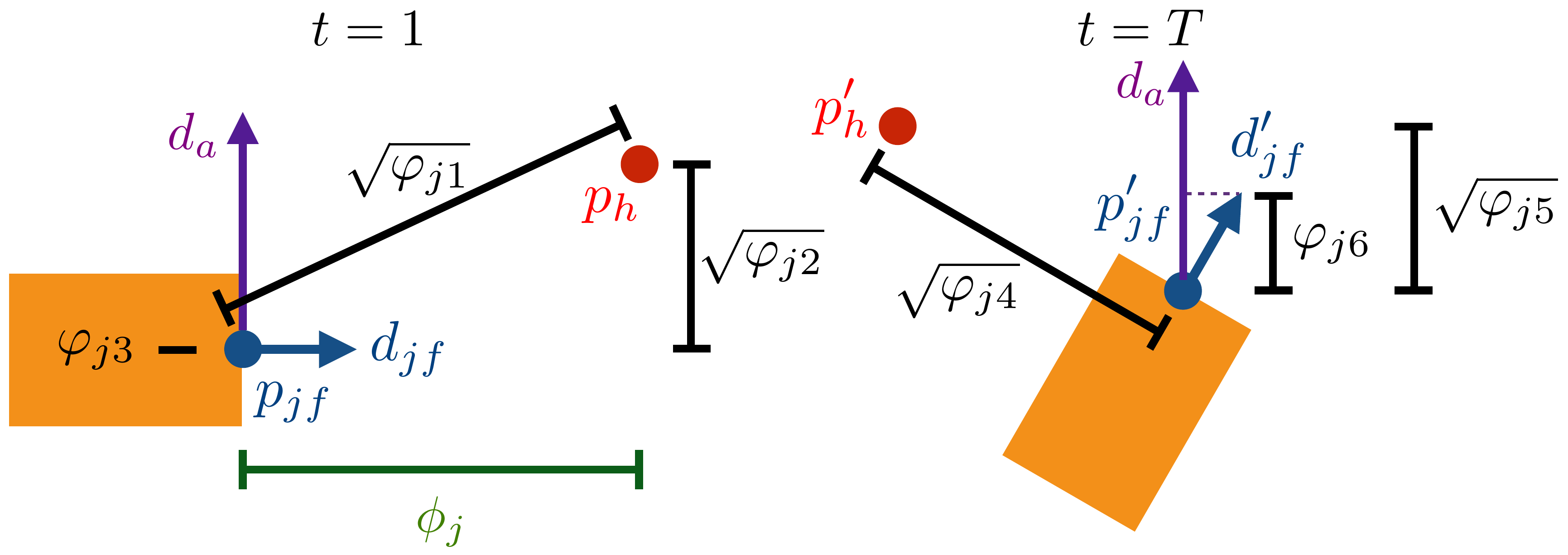}
\par\end{centering}

\protect\caption{\label{fig:meta-feats}This figure illustrates the meta features (black)
computed for a feature $\phi_{j}$. This feature (green) describes
the initial horizontal distance between the side part of the object
(orange) and the robot's hand $p_{h}$ (red). The feature is characterized
by a feature position $p_{jf}$ and a direction $d_{jf}$ shown in
blue. The prior is being computed for the movement in the vertical
action direction $d_{a}$ (purple). The left side of the figure shows
the initial object configuration $t=1$, while the right side shows
the final configuration $t=T$. The black lines indicate the meta
features. The seventh meta feature  $\varphi_{j7}\in\{-1,1\}$   (not shown) is set to $\varphi_{j7}=1$ 
as the feature defines the position, and not the size, of a part.}
\end{figure}
The next step is to generate a set of features describing the constellation
of objects and parts in a scene. A subset of these features will ultimately
be used by the robot to generalize the learned manipulation skill
to different situations. Rather than defining specific features for
one task, we want to define a generic set of rules for generating features
for arbitrary objects across different manipulation tasks. In order
to create a consistent set of features between different scenes, we
assume that the number of objects and the number of respective parts
are the same across different instances of a given task. 

To describe the objects in the scene, the robot needs features that
capture the locations and shapes of the individual object parts. The
positions and sizes of the parts are computed by first fitting an
axis-aligned bounding box to each of the parts. The x-y-z position
of the part is then given by the center of the bounding box. The position
is defined relative to the initial position of the robot's hand. The
x-y-z dimensions of the box are used as features to capture the size
of the part. This bounding box representation is relatively coarse,
and some of the geometric details of the parts are lost. However,
most of the extracted parts will have basic shapes. By using a part-based
approach, the proposed representation automatically creates more features
to describe the task-relevant regions of the objects. In the future,
the set of features could be extended to include other potentially relevant object properties,
e.g., the principal axes of the parts and the objects' masses.

Each task will have its own set of object features due to the automatic
feature generation. Hence, the robot needs a suitable method for computing
the similarity between features in order to transfer the feature-relevance
prior between tasks. We can define this similarity by using meta features
$\varphi_{jh}\forall h\in\{1,...,H\}$ to describe the features $\phi_{j}\forall j\in\{1,...,M\}$
as well as their relation to the skill component being learned \cite{Lee_ICML_2007}. 

We describe each feature $\phi_{j}$ by first associating it with
a 3D position $\vec{p}_{jf}$ and direction $\vec{d}_{jf}$. The position
$\vec{p}_{jf}$ corresponds to the initial position of the feature's
part. The direction $\vec{d}_{jf}$ corresponds to the direction in
which the feature is computed, e.g., $\vec{d}_{jf}=[1\:0\:0]^{T}$
if the feature $\phi_{j}$ describes the position or size of a part
along the x axis. In order to relate the feature to the skill learning,
we also define the initial 3D position of the hand $\vec{p}_{h}$
and the direction $\vec{d}_{a}$ of the skill component being learned,
i.e., $\vec{d}_{a}=[0\:0\:1]^{T}$ when learning the z component
of the manipulation skill. 

Given these variables, the first meta feature describes the initial
proximity between the feature and the robot's hand $\varphi_{j1}=||p_{jf}-p_{h}||^{2}$.
The second meta feature describes the proximity in the direction of
the skill component $\varphi_{j2}=||\vec{d}_{a}^{T}(p_{jf}-p_{h})||^{2}$,
and the third meta feature describes the alignment between the feature
and the skill component $\varphi_{j3}=|\vec{d}_{a}^{T}\vec{d}_{jf}|$.
The next three meta features, $\varphi_{j4}$ to $\varphi_{j6}$,
are computed in the same manner as the first three, but using the
feature position $\vec{p}'_{jf}$, feature direction $\vec{d}'_{jf}$,
and hand position $\vec{p}'_{h}$ at the end of the skill demonstration.
The skill direction $\vec{d}_{a}$ is kept constant while the feature
direction $\vec{d}_{jf}$ is assumed to move together with object
part, as illustrated in Fig. \ref{fig:meta-feats}. The seventh meta
feature $\varphi_{j7}$ indicates whether the feature represents the
position $\varphi_{j7}=1$ or the size $\varphi_{j7}=-1$ of a part.
The final meta feature $\varphi_{j8}=1$ is a bias term. 

Additional meta features could be explored in the future, e.g., the
feature's alignment with the part's mean normal, or if the part is
currently interacting with another part. 
Meta features could also be used to describe other types of object features, e.g., features generated by a convolutional neural network \cite{Levine_arxiv_2015}.
The current set of  meta
features allows the robot to capture the proximity and alignment of
the various object features. The meta features also adapt to the direction
of the skill component being learned and which hand is being used.
In Section \ref{sub:Probablisic-Model-for}, we explain how these
meta features can be used to compute a prior over the relevance of
the feature.

\section{Learning Relevant Features for Generalizing Manipulation Skills\label{sec:Learning-Relevant-Features}}

In order to perform manipulations under different situations, the
robot must learn a manipulation skill that adapts to the extracted
object features. We use dynamic motor primitives (DMPS) to represent
the skills. Our reformulation of the parameterized DMPs with object
features is explained in Section \ref{sub:Motor-Primitive-Skill}.
In order to select a relevant set of features for generalizing the
skill, the robot uses stochastic search variable selection (SSVS)
with a meta-level prior to transfer  information regarding feature relevance
between skills. The probabilistic model over the skill parameters is explained
in Section \ref{sub:Probablisic-Model-for}. The feature selection
process, based on Gibbs sampling, is described in Section \ref{sub:Inferring-Relevant-Object}.

\subsection{Motor Primitive Skill Representation\label{sub:Motor-Primitive-Skill}}

The robot's skills are represented using dynamic motor primitives
(DMPs) \cite{Ijspeert_ICRA_2002,Schaal_ISRR_2004}. A DMP consists
of linear dynamical systems of the form 
\begin{eqnarray*}
\ddot{y} & = & \alpha_{z}(\beta_{z}\tau^{-2}(y_{0}-y)-\tau^{-1}\dot{y})+\tau^{-2}\sum_{j=1}^{M}\phi_{j}f(x;\vec{w}_{j}),
\end{eqnarray*}
where $y$ is the state, $y_{0}$ is the initial state, $\alpha_{z}$
and $\beta_{z}$ are constants that define the spring and damper coefficients,
$\tau$ is a time coefficient, $x$ is the state of the canonical
system, and $f$ is a forcing function. Each skill component, e.g.,
the hand's movement in the $z$ direction, is modeled by a separate
linear system. The canonical state $x$ acts as a timer for synchronizing
multiple linear systems. It starts at $x=1$ and decays according
to $\dot{x}=-\tau x$. In our framework, the amplitude parameters
$\phi_{i}$ correspond to the object features in order to allow the skill
to adapt to different scenarios. The adaptation of the skill's trajectory
to an object feature $\phi_{j}$ is defined by its corresponding forcing
function $f(x;\vec{w}_{j})$, where the $k$\textsuperscript{th} element of the vector
$\vec{w}_{j}\in\mathbb{R}^{K}$ is given by the parameter $[\vec{w}_{j}]_{k}=w_{jk-1}$.
The forcing function $f$ is represented using a locally weighted
regression of the form 
\[
f(x;\vec{w}_{j})=\alpha_{z}\beta_{z}\left(\frac{\sum_{k=1}^{K}\psi_{k}(x)w_{jk}x}{\sum_{k=1}^{K}\psi_{k}(x)}+w_{i0}\psi_{0}(x)\right),
\]
where $\psi_{k}\forall k\in\{1,...,K\}$ are Gaussian basis functions,
and $\psi_{0}$ is a basis function that follows a minimum jerk trajectory
from $0$ to $1$. The shape of the DMP's trajectory is defined by
the parameters $w_{ij}$ together with their corresponding basis functions.
The standard formulation of the DMPs, with the explicit goal-state
parameter $g$, can be obtained by setting $w_{1k}=0\forall k\in\{1,...K\}$,
$w_{10}=1$, $\psi_{0}(x)=1$, and $\phi_{1}=g-y_{0}$. In our reformulation,
the first term of the linear system is defined entirely by the initial
state of the robot and its environment, and the the goal states are
absorbed into the object features $\phi_{i}$.

\subsection{Probabilistic Model for Skill Learning with Sparse Features\label{sub:Probablisic-Model-for}}

\begin{figure}
\centering{}\includegraphics[width=0.8\columnwidth]{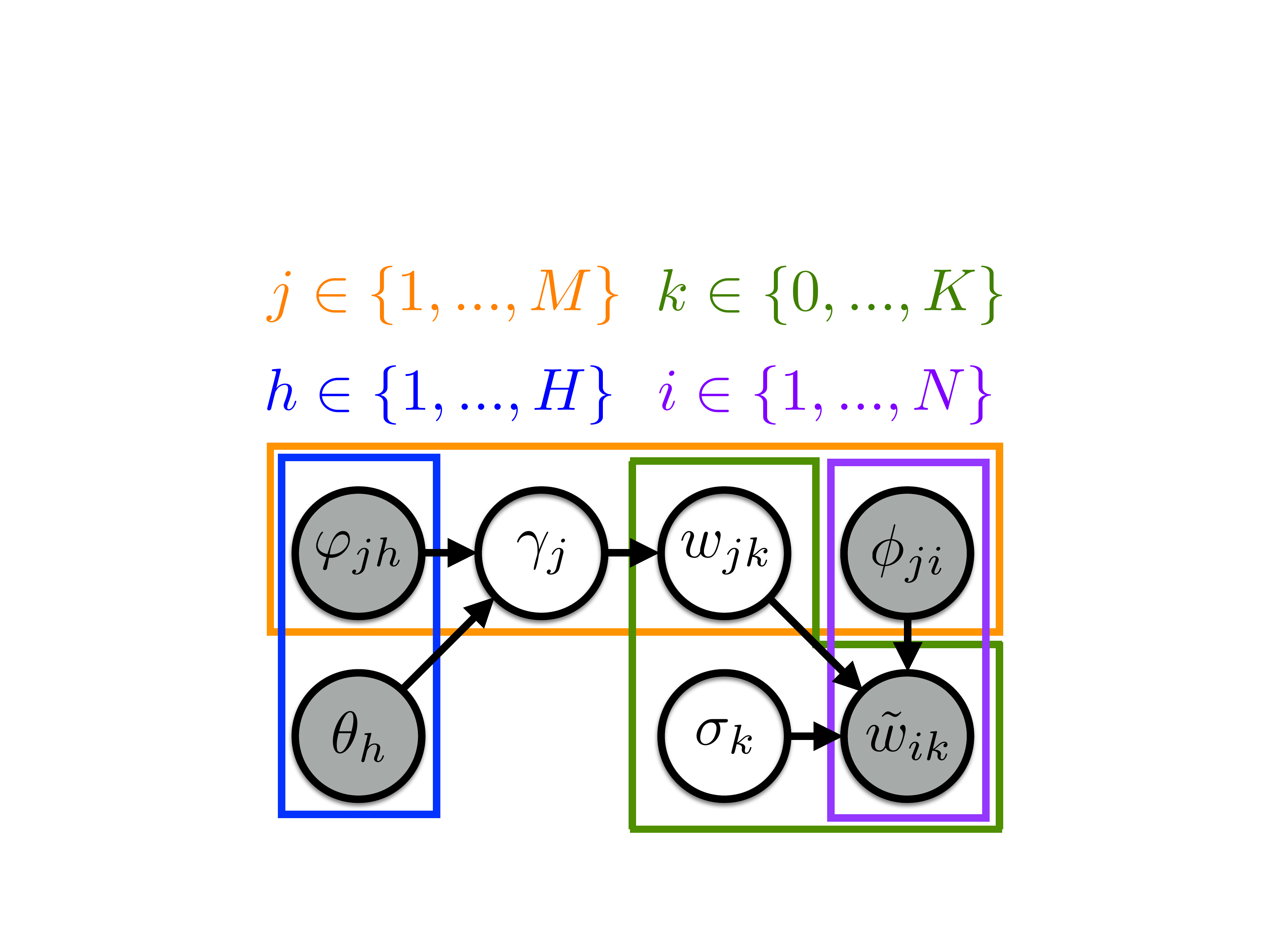}\protect\caption{\label{fig:The-graphical-model}The graphical model for the skill
parameters. Shaded nodes indicate observed variables, and white nodes
represent latent variables. Repeated variables are indicated by the
square plates and the variables' indices. The figure is best viewed
in color. }
\end{figure}
The next step is to learn the shape parameters $w_{jk}$, which determine
how the DMP adapts to changes in the object features. In theory, the
robot could learn a skill that adapts to every object feature extracted
from the scenes. However, in practice, most of the object features
will be irrelevant for generalizing the skill and the corresponding
skill parameters should be close to zero. The robot should therefore learn
to select the relevant object features and ignore the irrelevant ones. 

Our approach to learning the relevant object features is based on
the stochastic search variable selection (SSVS) approach \cite{George_ASA_1993}.
The graphical model of our framework is shown in Fig. \ref{fig:The-graphical-model}.
As the relevant object features may be different for each linear system
of the DMP, we perform a separate feature selection for each skill
component. This probabilistic model captures not only the distribution
over the skill parameters $\vec{w}_{j}$, but also the relevance $\gamma_{j}\in\{0,1\}$
of the object feature $\phi_{j}$. 

The relevance $\gamma_{j}$ is distributed according to a Bernoulli
distribution $\gamma_{j}\sim\text{Bern}((1+\exp(-\vec{\varphi_{j}}^{T}\vec{\theta}))^{-1})$
where the meta features $\vec{\varphi}_{j}\in\mathbb{R}^{H}$ and
the hyperparameters $\vec{\theta}\in\mathbb{R}^{H}$ define the prior
distribution over the feature's relevance. The hyperparameters $\vec{\theta}$
are known as the \emph{meta prior} \cite{Lee_ICML_2007}. Since the prior
on $\gamma_{j}$ has a logistic regression form, the values of the
meta prior are learned from previous skills using iteratively reweighted
least squares.

The relevance $\gamma_{j}$ determines the prior variance over the
shape parameters $\vec{w}_{j}$ for the $j$\textsuperscript{th}
object feature $\phi_{j}$. If the feature is relevant $\gamma_{j}=1$,
then the distribution over the shape parameters $w_{jk}\forall k\in\{0,...,K\}$
is given by $p(w_{jk}|\gamma_{j}=0,\check{s},\hat{s})=\mathcal{N}(0,\check{s}^{2})$,
where the standard deviation $\check{s}$ defines a broad Gaussian.
If the feature is irrelvant $\gamma_{j}=0$, then the distribution
is given by $p(w_{jk}|\gamma_{j}=1,\check{s},\hat{s})=\mathcal{N}(0,\hat{s}^{2})$,
where $\hat{s}\ll\check{s}$. In this manner, the shape parameters
corresponding to an irrelevant feature are more likely to be close
to zero and, hence, have less effect on the generalization of the
skill. 

The distribution over the skill parameters is to be inferred from
a set of $N$ training demonstrations. The $i$\textsuperscript{th }
training demonstration includes a set of values for the object features
$\phi_{ji}\forall j\in\{1,..M\}$ and a trajectory $\xi_{i}$ representing
the robot's actions. Rather than working directly with the trajectory
data $\xi_{i}$, we transform each trajectory into a separate set
of target shape parameters $\tilde{w}_{ik}$. These target parameters
are equivalent to the shape parameters if the DMP were learned using
only a single constant object feature $\phi_{1}=1$. The goal is thus
to learn a set of skill parameters $\vec{w}_{j}$ that approximate
these target values $\tilde{w}_{ik}\approx\sum_{j=1}^{M}w_{jk}\phi_{ji}$
for all of the training trajectories. We model the distribution over
the target values using a normal distributions $\tilde{w}_{ik}\sim\mathcal{N}(\sum_{j=1}^{M}w_{jk}\phi_{ji},\sigma_{k}^{2})$,
where $\sigma_{k}^{2}$ is the output variance in the shape parameters.
We model the distribution over these variances using an inverse-gamma
distribution $\sigma_{k}^{2}\sim\text{Inv-Gamma}(a,b)$. We set the
shape and scale parameters $a$ and $b$ to constants, which
worked well in the experiments. Alternatively, the hyperparameters
can depend on the number of relevant features $\sum_{j}\vec{\gamma}_{j}$
such that the variance is smaller if the model uses more features
\cite{George_ASA_1993}.

\subsection{Inferring Relevant Object Features\label{sub:Inferring-Relevant-Object}}

Given a matrix $\tilde{\vec{W}}\in\mathbb{R}^{N\times M}$ of target
shape parameters $[\tilde{\vec{W}}]_{i,j}=\tilde{w}_{ij}$ and a matrix
$\vec{\Phi}\in\mathbb{R}^{M\times N}$ of object features $[\vec{\Phi}]_{j,i}=\phi_{ji}$
from $N$ demonstrations, the next step is to estimate the relevant
features $\vec{\gamma}$ for generalizing the manipulation skill to
new scenarios. In particular, we want to determine which feature are
likely to be relevant $\gamma_{j}=1$ under the posterior distribution
$p(\vec{\gamma}|\tilde{\vec{W}},\vec{\Phi})$. The posterior distribution
$p(\vec{\gamma}|\tilde{\vec{W}},\vec{\Phi})$ is given by

\[
p(\vec{\gamma}|\tilde{\vec{W}},\vec{\Phi})=\int\int p(\vec{\gamma},\vec{W},\vec{\sigma}|\tilde{\vec{W}},\vec{\Phi})\text{d}\vec{W}\text{d}\vec{\sigma}
\]
where $\vec{W}\in\mathbb{R}^{M\times K+1}$ is a matrix of all of
the shape parameters $[\vec{W}]_{j,k}=w_{jk}$, and $\vec{\sigma}$
is the set of all standard deviations $\sigma_{k}$ for the target
weights. Although this distribution cannot be computed in closed form,
it can be efficiently approximated using a Gibbs sampling approach
\cite{George_ASA_1993,Geman_PAMI_1984}. Gibbs sampling is a Markov chain Monte Carlo
method that allows us to draw samples from the joint distribution
$p(\vec{\gamma},\vec{W},\vec{\sigma}|\tilde{\vec{W}},\vec{\Phi})$
by iteratively sampling over each of the individual components, i.e.,
$\vec{\gamma}$, $\vec{W}$, and $\vec{\sigma}$, given all of the
other components. 

\begin{figure}
\begin{centering}
\includegraphics[width=0.9\columnwidth]{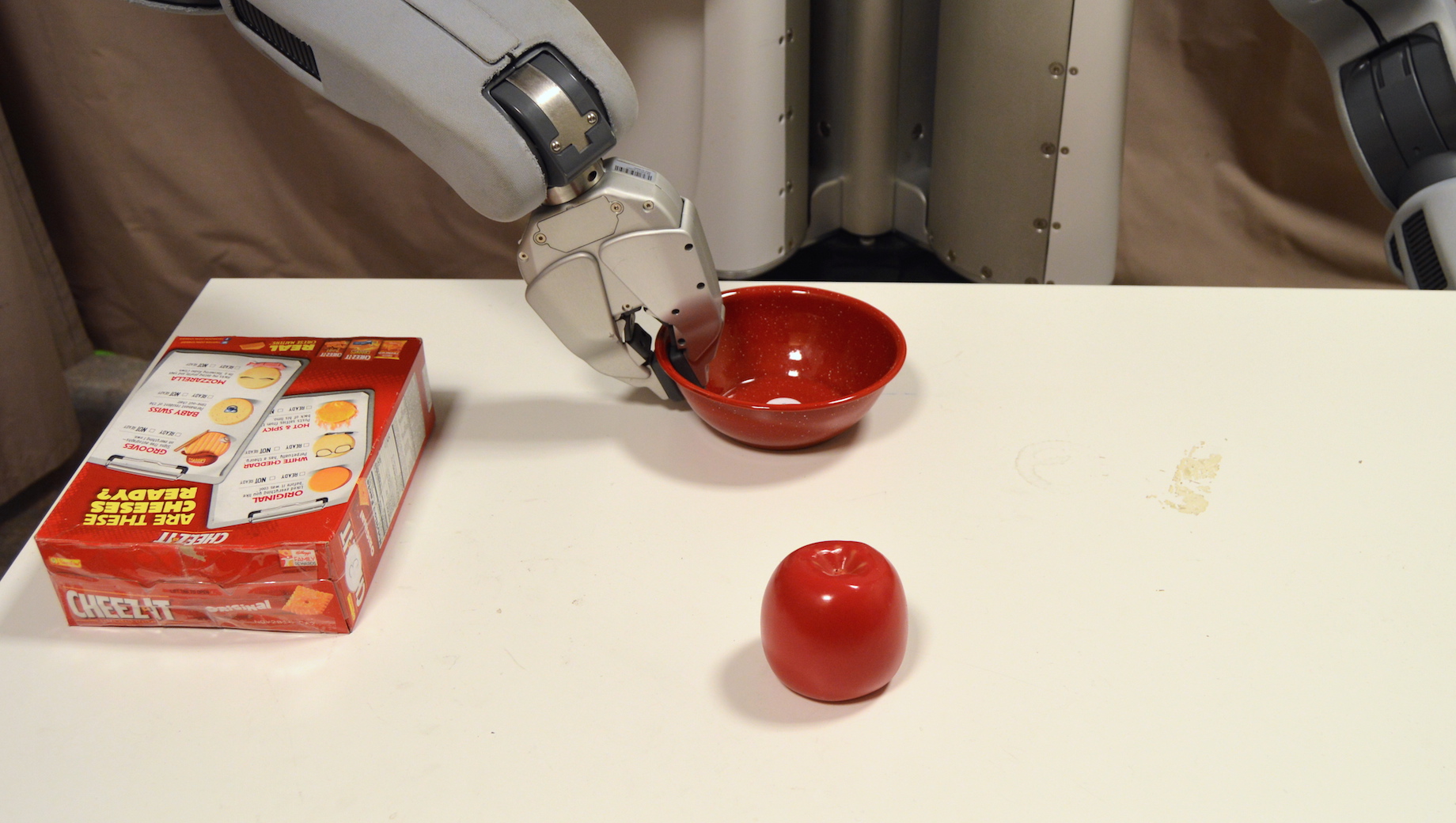}
\includegraphics[width=0.9\columnwidth]{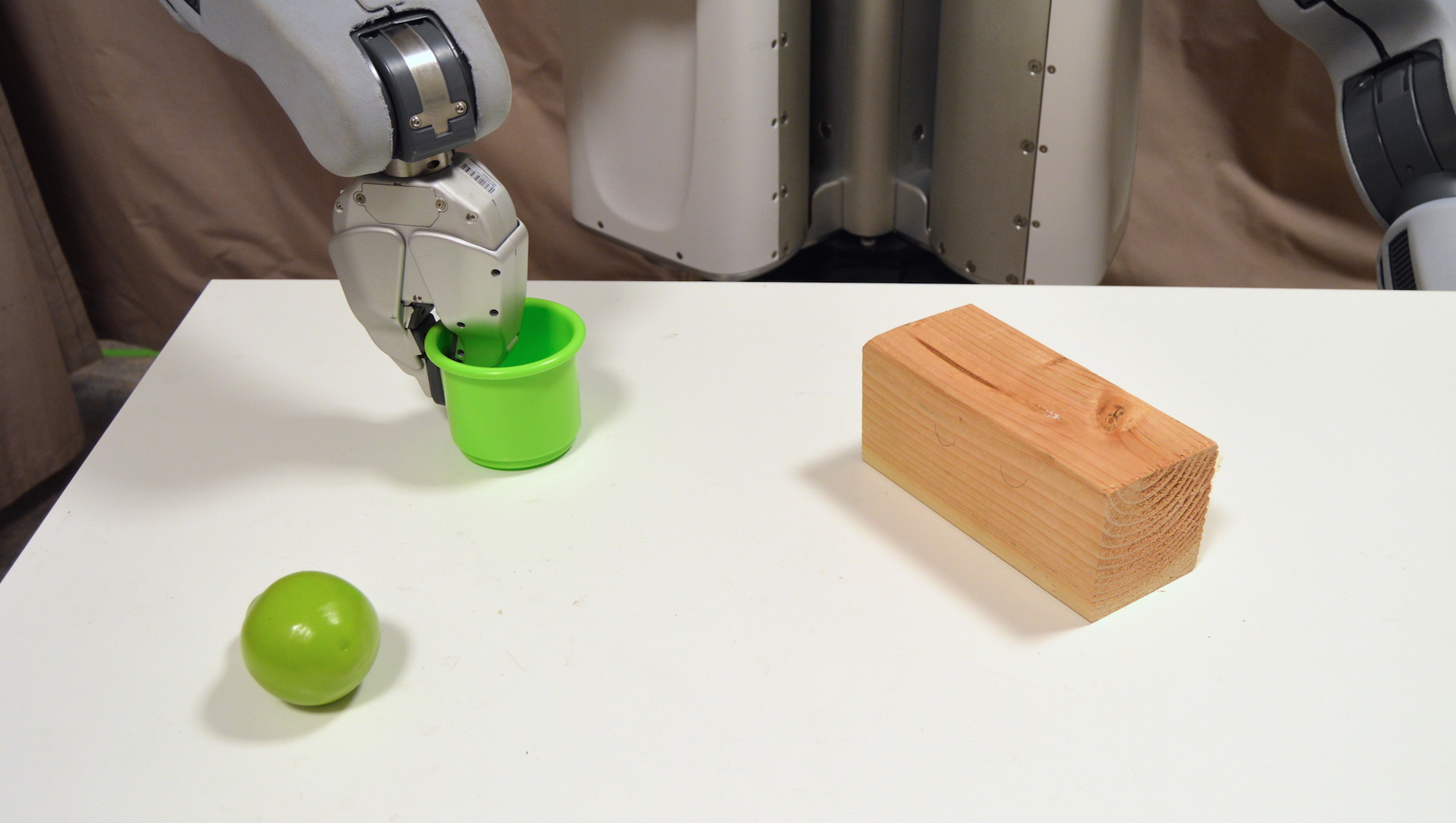}
\includegraphics[width=0.9\columnwidth]{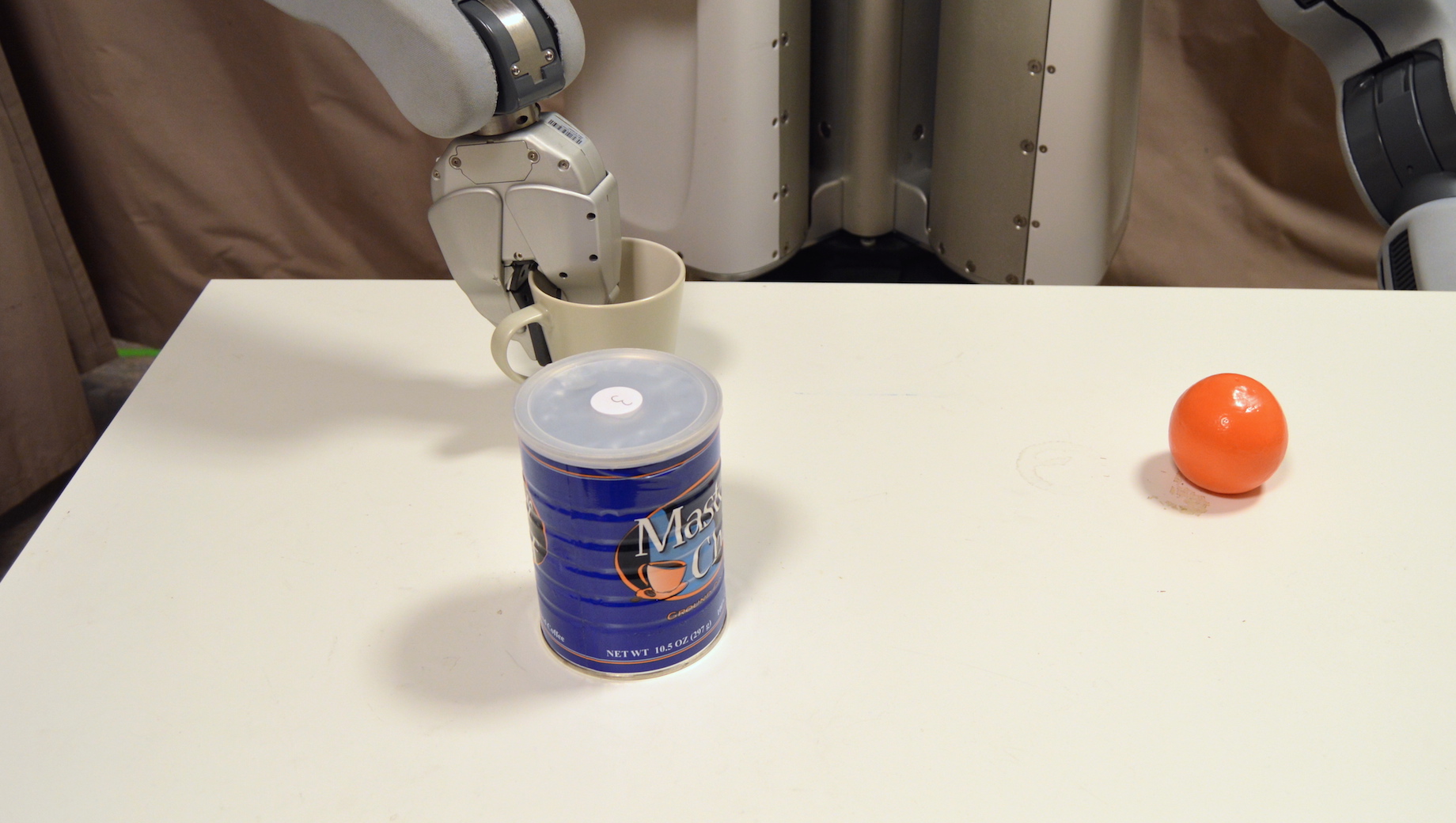}
\par\end{centering}
\protect\caption{ \label{fig:exampScenes} Example scenes for the  placing task. The task is to place the grasped object centered on the box, block,  or can depending on the scene.}
\end{figure}

We initially set all of the relevance parameters $\vec{\gamma}$ to
one and the standard deviations $\vec{\sigma}$ to a predefined value.
We also compute the priors over each relevance parameter $p(\gamma_{j})$
based on their respective meta features $\varphi_{jh}$ and the meta prior $\theta_{h}$.
We then sample a new set of shape parameters $\vec{w}_{k}\in\mathbb{R}^{M}\forall k\in\{0,...,K\}$
according to 
\[
\vec{w}_{k}\sim\mathcal{N}\left((\vec{\Phi}\vec{\Phi}^{T}+\sigma_{k}^{2}\vec{R}^{-1})^{-1}\vec{\Phi}\tilde{\vec{w}}_{k}|(\vec{R}^{-1}+\sigma_{k}^{-2}\vec{\Phi}\vec{\Phi}^{T})^{-1}\right)
\]
where $\tilde{\vec{w}}_{k}\in\mathbb{R}^{N}$ is a vector containing
the $N$ samples' target shape parameters for the $k$\textsuperscript{th} basis function
of the DMP, and $\vec{R}\in\mathbb{R}^{M\times M}$ is a diagonal
matrix with elements $[\vec{R}]_{jj}=\hat{s}^{2}$ if $\gamma_{j}=0$
and $[\vec{R}]_{jj}=\check{s}^{2}$ if $\gamma_{j}=1$. Given these
updated shape parameters, we then sample a new set of variance terms
$\sigma_{k}^{2}\forall k\in\{0,...K\}$ using an inverse gamma distribution
\[
\sigma_{k}^{2}\sim\text{Inv-Gamma}(a,b+(2(\tilde{\vec{w}}_{k}-\vec{\Phi}^{T}\vec{w}_{k})(\tilde{\vec{w}}_{k}-\vec{\Phi}^{T}\vec{w}_{k}))^{-1}).
\]
Finally, using the updated shape and variance parameters, we sample
a new set of relevance parameters $\gamma_{j}$ according to the Bernoulli
distribution
\[
\gamma_{j}\sim\text{Bern}\left(Z_{j}^{-1}\mathcal{N}(\vec{w}_{j}|0,\check{s}^{2}I)p(\gamma_{j}=1)\right),
\]
where 
\[
Z_{j}=\mathcal{N}(\vec{w}_{j}|0,\hat{s}^{2}I)p(\gamma_{j}=0)+\mathcal{N}(\vec{w}_{j}|0,\check{s}^{2}I)p(\gamma_{j}=1),
\]
and the meta prior $\vec{\theta}$ and meta features $\vec{\varphi_{j}}$
are used to compute the prior 
\[
p(\gamma_{j}=1)=1-p(\gamma_{j}=0)=(1+\exp(-\vec{\varphi_{j}}^{T}\vec{\theta}))^{-1}.
\]
This sampling process continues to iterate between the different components
to generate more samples from the posterior. 

The Gibbs sampling provides the robot with a distribution over the
relevance parameters. In order to ultimately select a set of relevant
features, the robot computes the maximum a posteriori (MAP) estimate
of the the relevance parameters. Hence, the robot selects a feature
$\phi_{j}$ to be relevant iff the majority of the samples from the
Gibbs sampling were $\gamma_{j}=1$  \cite{George_ASA_1993}. This MAP estimate also corresponds
to the Bayes estimate of the relevance parameters under a $0/1$ loss
function. The skill parameters $\vec{w}_j$ for the selected features are then computed 
using linear regression. When presented with a new scenarios,  the robot extracts
the object features and uses the learned skill parameters to perform the manipulation task.

\section{Experiments\label{sec:Experiments}}

\begin{figure}[t]
\begin{centering}
\includegraphics[width=1\columnwidth]{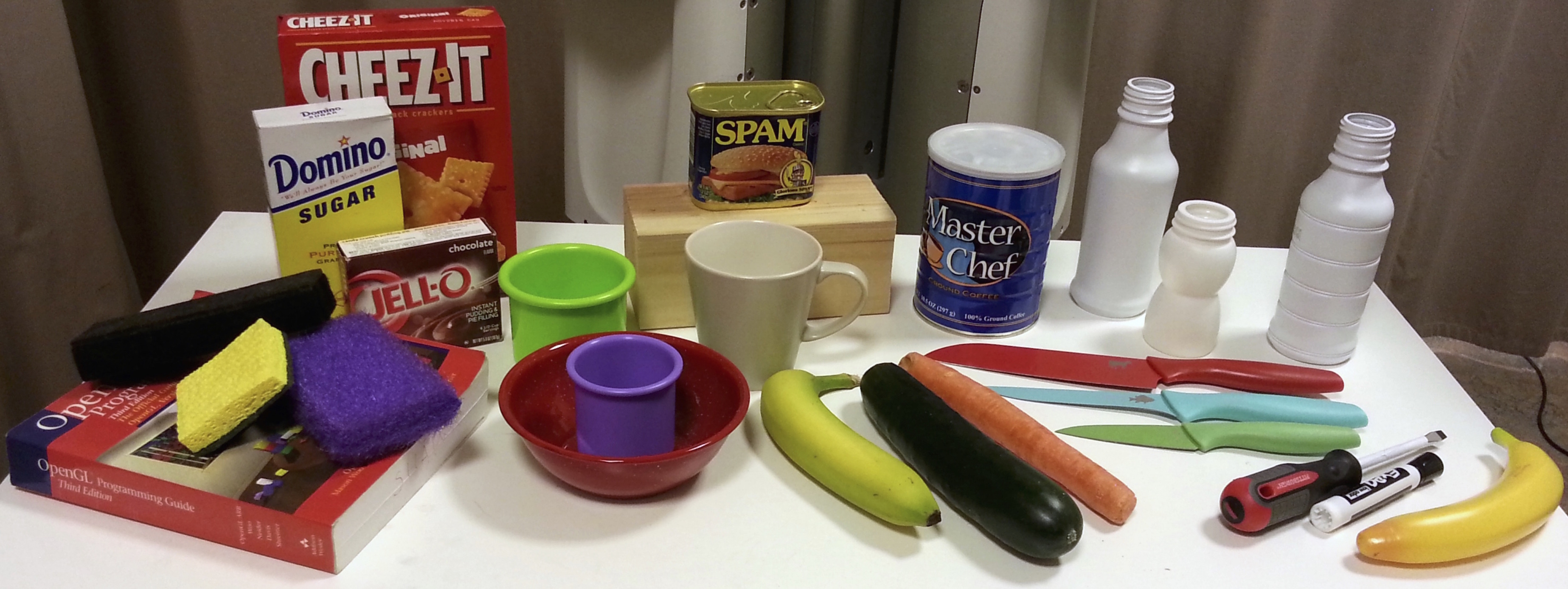}
\par\end{centering}
\protect\caption{\label{fig:objects}The image shows the manipulated objects from the experiments. Some of the objects were used for learning more than one skill}
\end{figure}

The proposed approach was evaluated on a series of manipulation tasks
using a PR2 robot, as shown in Fig. \ref{fig:overview}. The robot has
two arms with seven degrees of freedom each. Each arm is equipped
with a parallel-jaw gripper. The robot observes the scene using a
head-mounted kinect. The extraction of the object parts based on the
skill demonstrations is detailed in Section \ref{sub:Extracting-Object-Parts}.
In Section \ref{sub:Benchmarking-Priors-for}, we discuss a benchmark
experiment comparing the performance of the feature selection when
using a meta prior versus a standard uniform prior. In a second benchmark experiment,
we evaluated the accuracy of the predicted goal states of the learned skills, as described in Section \ref{sub:GoalPred}.
The evaluation of the robot's ability to learn skills using feature selection is discussed in 
Section \ref{sub:Learning-Manipulation-Skills}. 

\subsection{Extracting Object Parts\label{sub:Extracting-Object-Parts}}

\begin{figure*}
\begin{centering}
\begin{tabular}{c|c|c|c|c|c}
Place & Push & Tilt & Cut & Pour & Wipe\tabularnewline
\hline 
\includegraphics[scale=0.25]{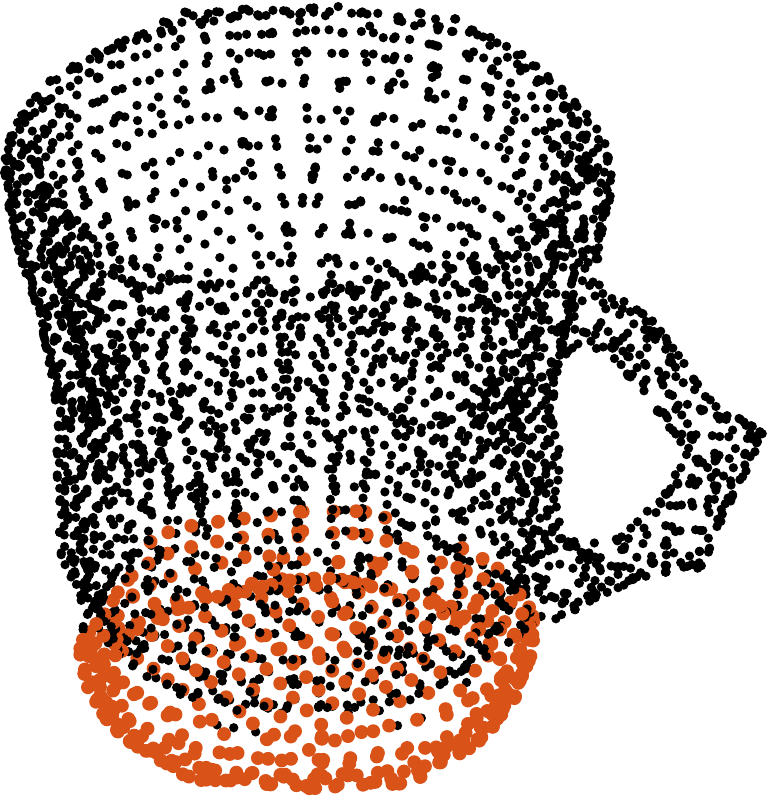} & \includegraphics[scale=0.3]{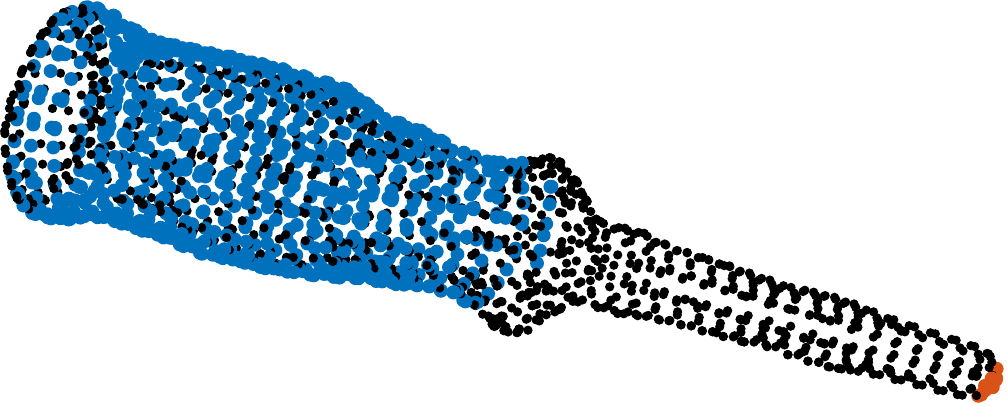} & \includegraphics[scale=0.35]{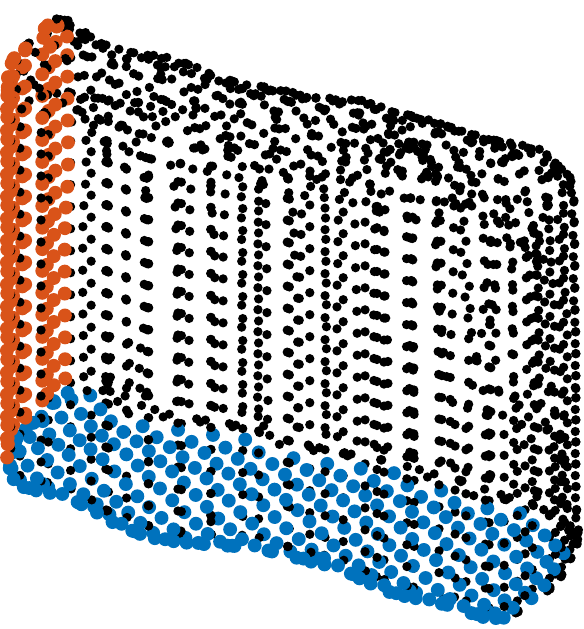} & \includegraphics[scale=0.32]{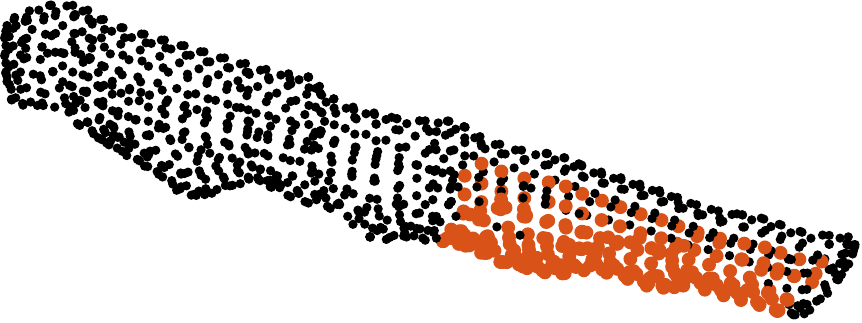} & \includegraphics[scale=0.24]{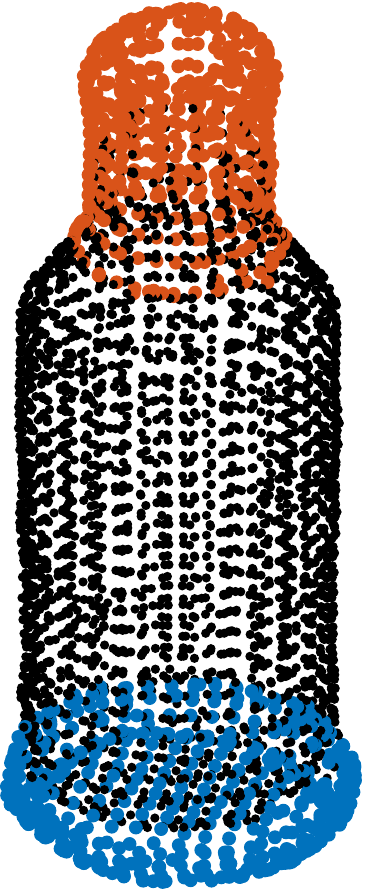} & \includegraphics[scale=0.23]{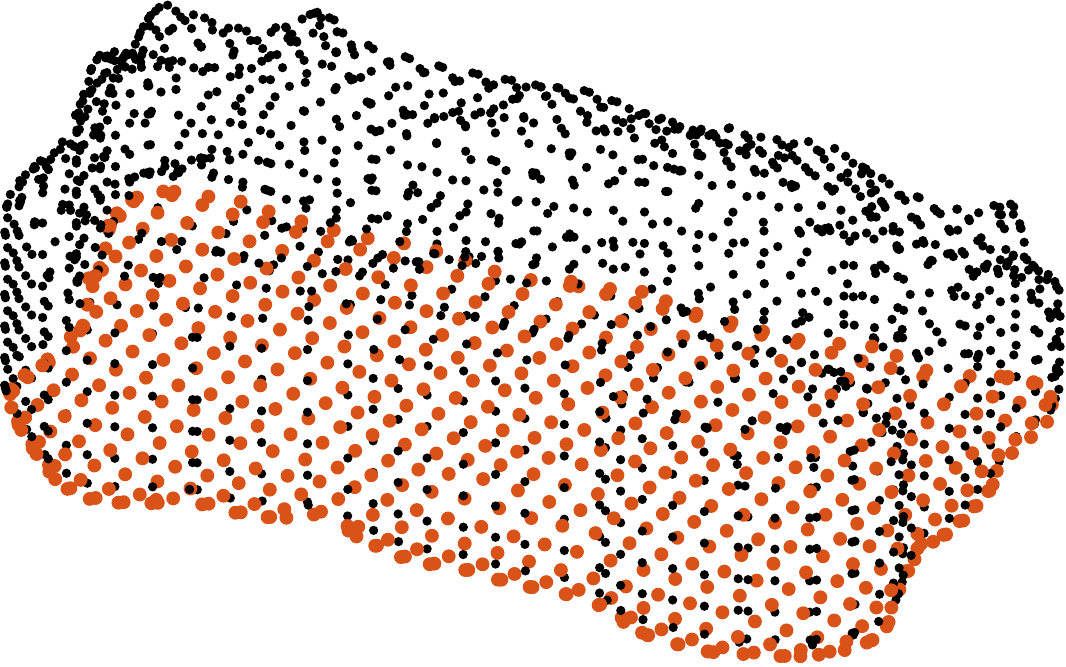}\tabularnewline
a) Cup & c) Screwdriver & e) Box & g) Knife & i) Bottle & k) Sponge\tabularnewline
\includegraphics[scale=0.25]{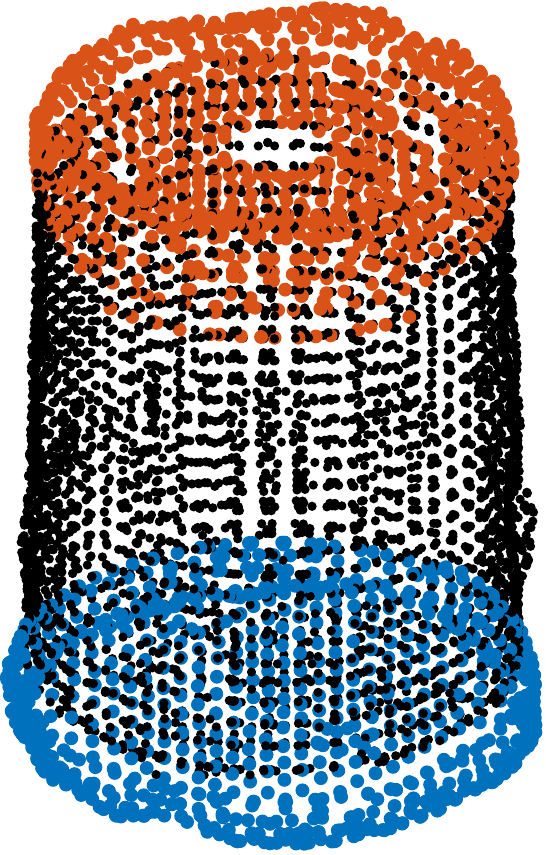} & \includegraphics[scale=0.25]{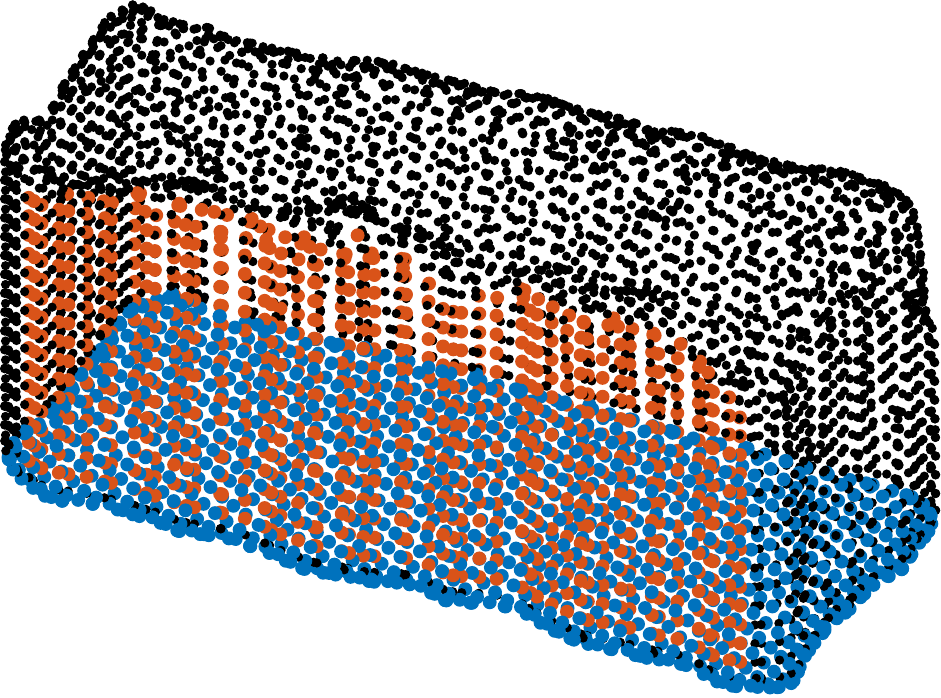} & \includegraphics[scale=0.35]{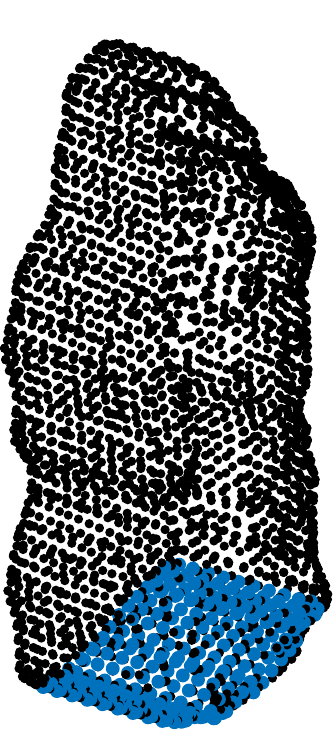} & \includegraphics[scale=0.25]{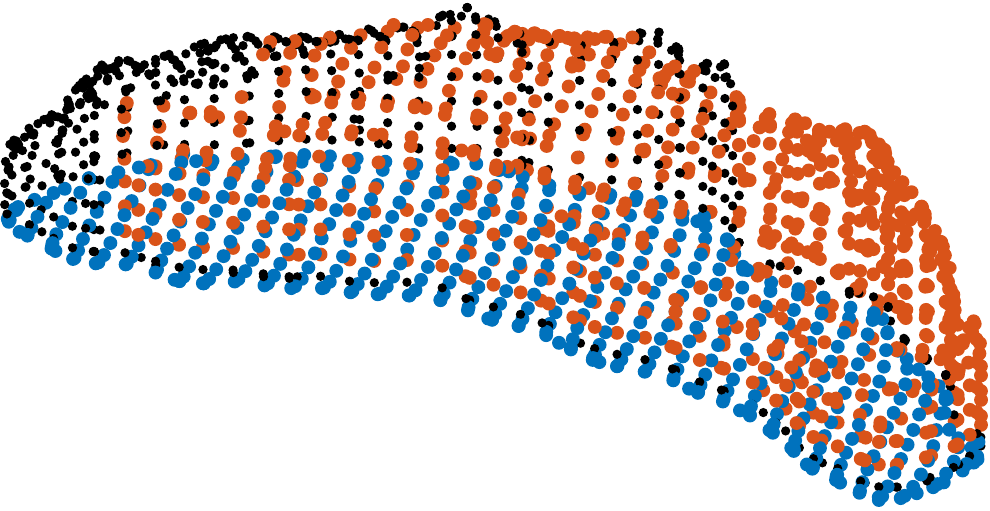} & \includegraphics[scale=0.25]{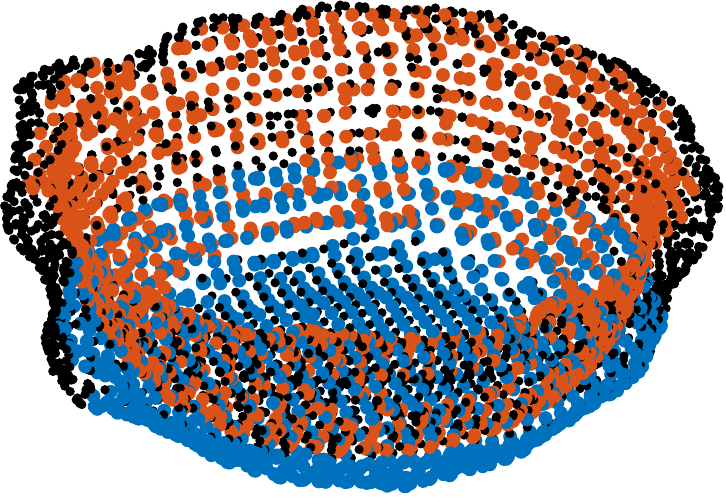} & \includegraphics[scale=0.25]{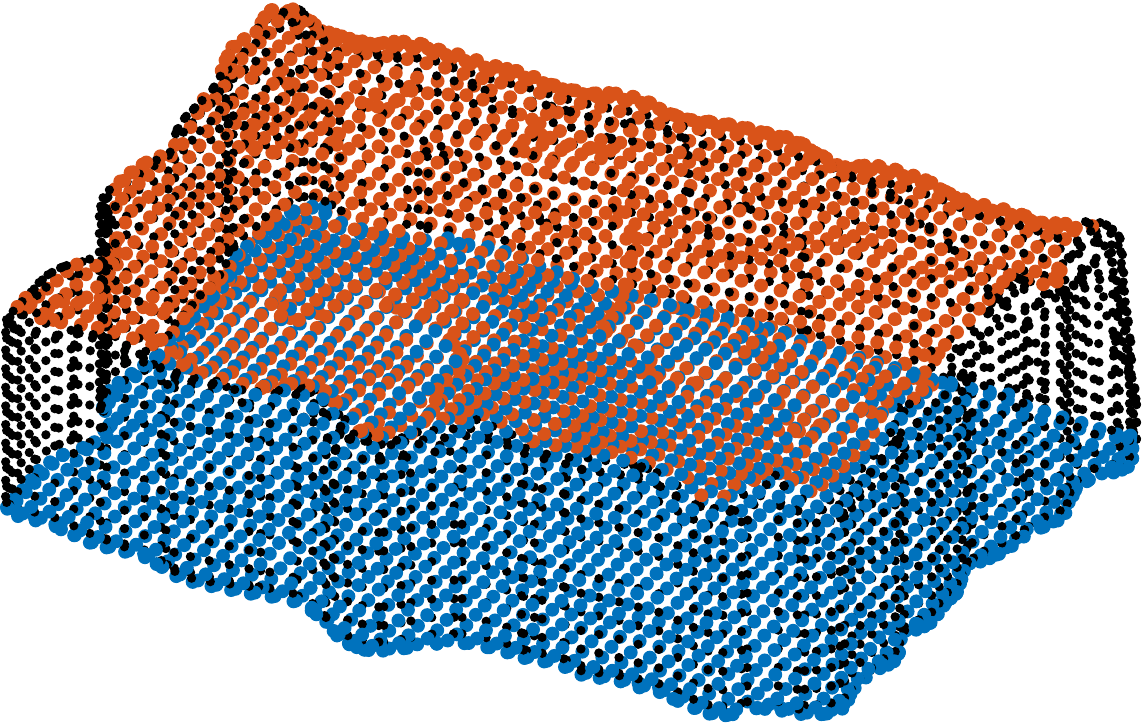}\tabularnewline
b) Large Can & d) Wooden Block & f) Bottle & h) Banana & j) Bowl & l) Box\tabularnewline
\end{tabular}
\par\end{centering}
\protect\caption{\label{fig:object-parts}Examples of object parts extracted from demonstrations
of manipulations. The points show the objects' 3D models. The
different colored points indicate parts of the point cloud. The top
row consists of the grasped objects, while the bottom row indicates
the objects that they interacted with using the red parts. For the
tilting task, the red part corresponds to the supporting surface at
the end of the task, and the mustard bottle (f) is irrelevant to the task. }
\end{figure*}

The robot was initially shown demonstrations, using kinaesthetic teaching,
of each of the following tasks: placing an object on top of another
object (place), pushing an object with an elongated object (push),
tilting a box between two adjacent sides (tilt), cutting an object
with a knife (cut), emptying a bottle into a container (pour), and
using an object to wipe the surface of another object (wipe). Each
task was demonstrated five times with three different sets of objects,
resulting in a total of $6\times5\times3=90$ demonstrations. 
The variety in the task scenarios is illustrated  in Fig. \ref{fig:exampScenes}, which shows
example scenes for the placing task. Most of the objects used in the 
experiment are from the YCB object set \cite{Bcalli_CRR_2015}. 

Each demonstration was performed with three objects in the scene,
although only one or two were relevant depending on the task. 
The manipulated objects are shown in Fig. \ref{fig:objects}. Coarse
3D models of the objects were computed from a single view
point by extruding their shapes. Objects that moved during the demonstrations
were tracked using the point cloud library's particle filter tracker
\cite{Fox_NIPS_2001,Rusu_ICRA_2011}. We assumed that grasped objects moved together
with the robot's hand in order to track them.

Given the set of demonstrations and object models, the robot extracted
the object parts as explained in Section \ref{sub:Detecting-Affordance-bearing-Parts}.
The GrabCut segmentation was performed using a coefficient of one
for the pairwise potentials, and mixtures of three Gaussians for the
unary potentials. Points in the point cloud model were considered
to be interacting with other objects if they were within $2$cm of
the other object's points and the inner product between their normals
was less than $-0.5$. The trajectories were subsampled at $1$Hz
in order to reduce the computation time. The spectral clustering was
performed using a Bhattacharyya kernel with the additional covariance hyperparameters set to $0.025$m
for the positions and $0.25$ for the normals \cite{Jebara_COLT_2003,Kroemer_IROS_2014}

Due to variations in the demonstrations and the stochasticity in the tracking
and segmentation, the demonstrations resulted in varying numbers of
object parts. Therefore, one demonstration that had the same number
of object parts as the mode was selected as a prototypical example
for each task. The parts of the other demonstrations were then matched
to these demonstrations by selecting the most similar part. The similarity
between parts was again computed using a Bhattacharyya kernel \cite{Jebara_COLT_2003,Kroemer_IROS_2014}.
In this manner, we created a consistent set of matched object parts
across the different demonstrations. 

From the demonstration's $270$ object instances, $87.4\%$ already
had the same number of parts as their respective modes, $11.1\%$
had more parts, and $1.5\%$ had less parts. Demonstrations with additional
parts were a minor issue as the matching process removed redundant
and erroneous parts. In one pushing demonstration, the object was
pushed close to the irrelevant object. This interaction generated
additional parts which were then removed by the matching process. 

Demonstrations with fewer parts are the result of interactions not
being detected due to poor demonstrations or tracking. For example,
while most of the pouring demonstrations were performed close to the
container's rim, one demonstration involved pouring from a
larger height of about $15$cm. This interaction was not detected
and hence no corresponding parts were generated. The interaction could 
be detected by observing the fluid \cite{Aksoy_IJRR_2011},
but currently only the containers are being tracked. For these demonstrations,
the matching process chose another part to replace the undetected
parts. 

\begin{figure*}
\begin{centering}
\includegraphics[width=0.99\columnwidth]{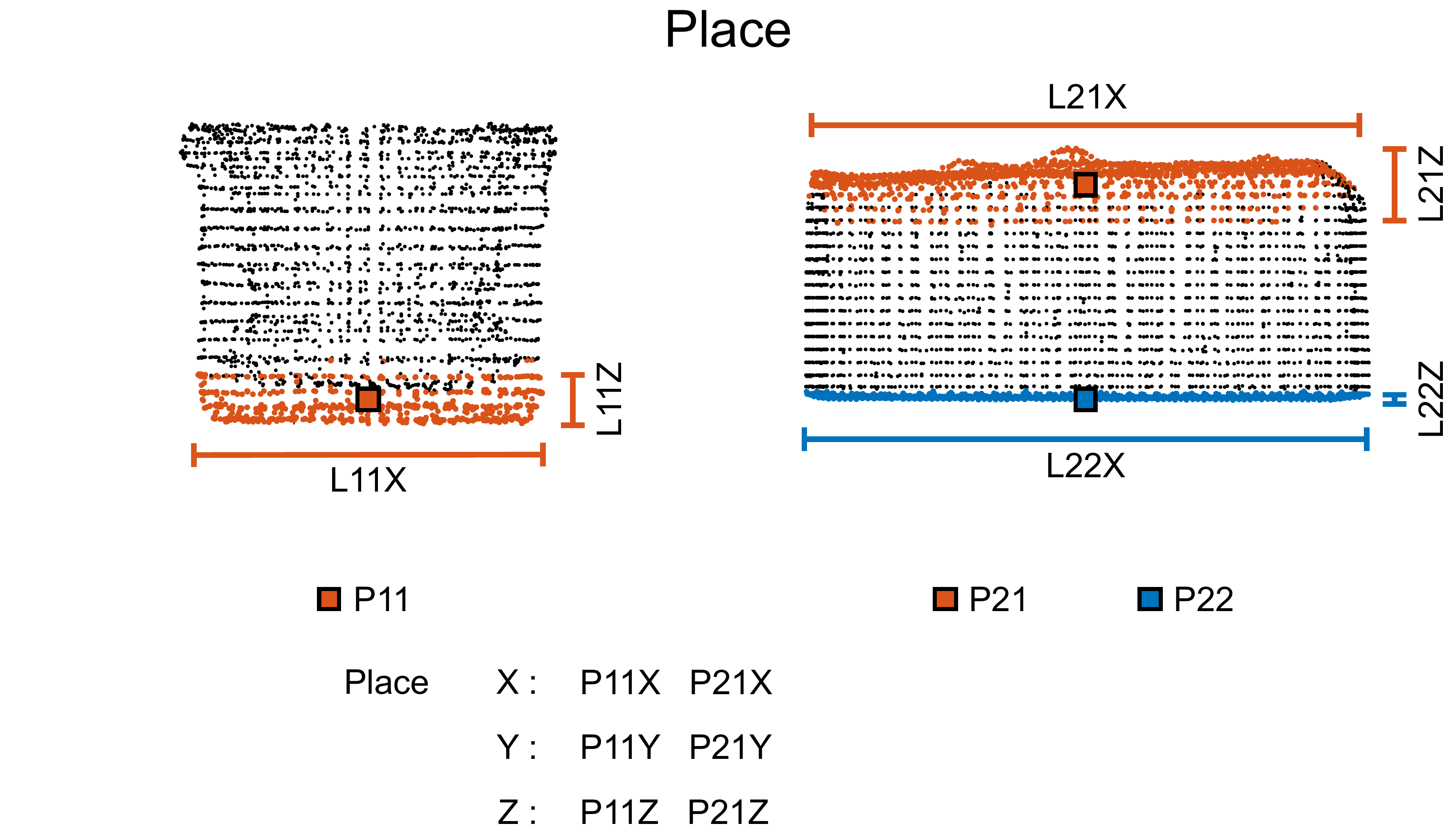}\hspace{0mm}
\includegraphics[width=0.99\columnwidth]{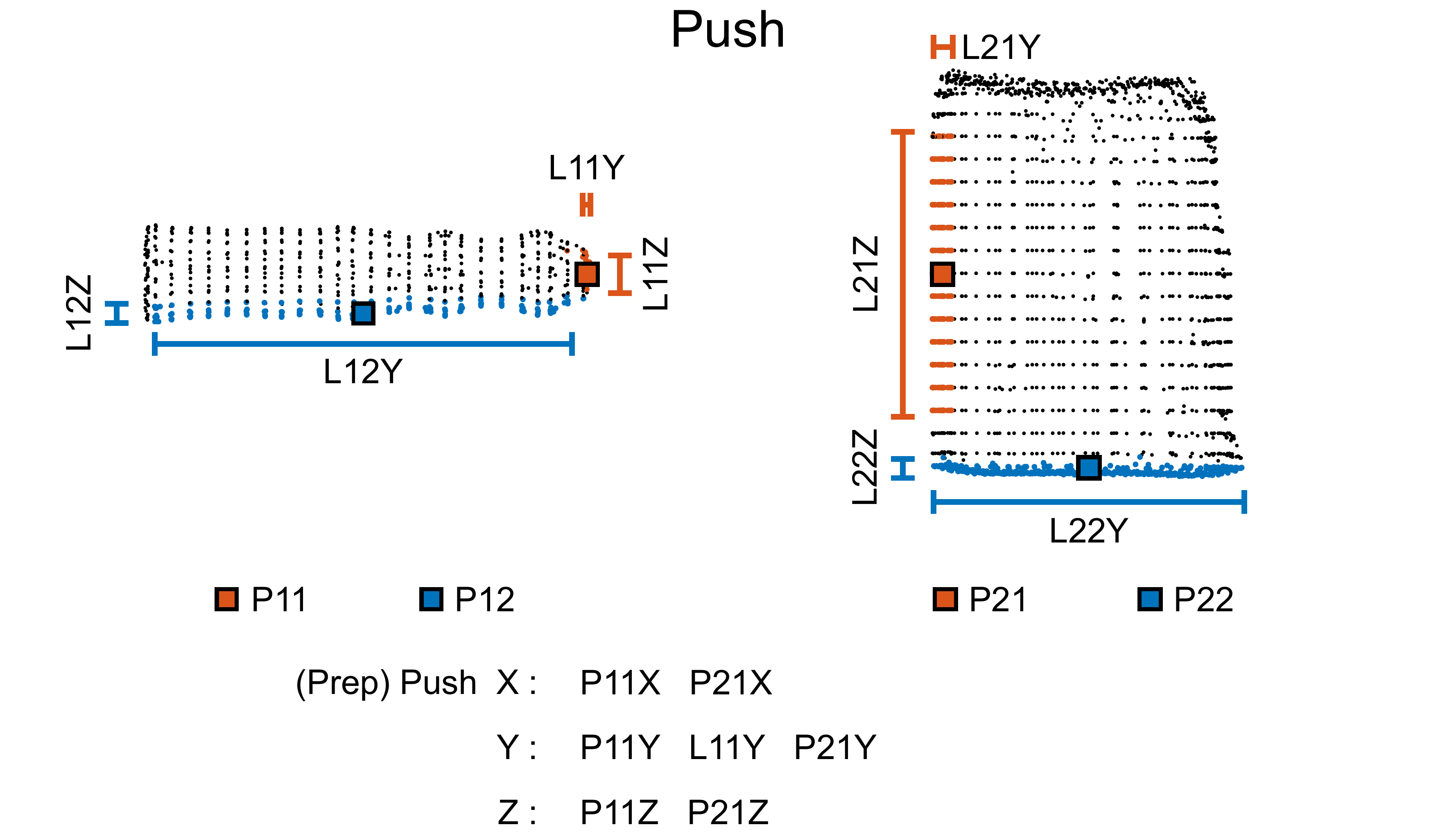}\vspace{10mm}
\includegraphics[width=0.99\columnwidth]{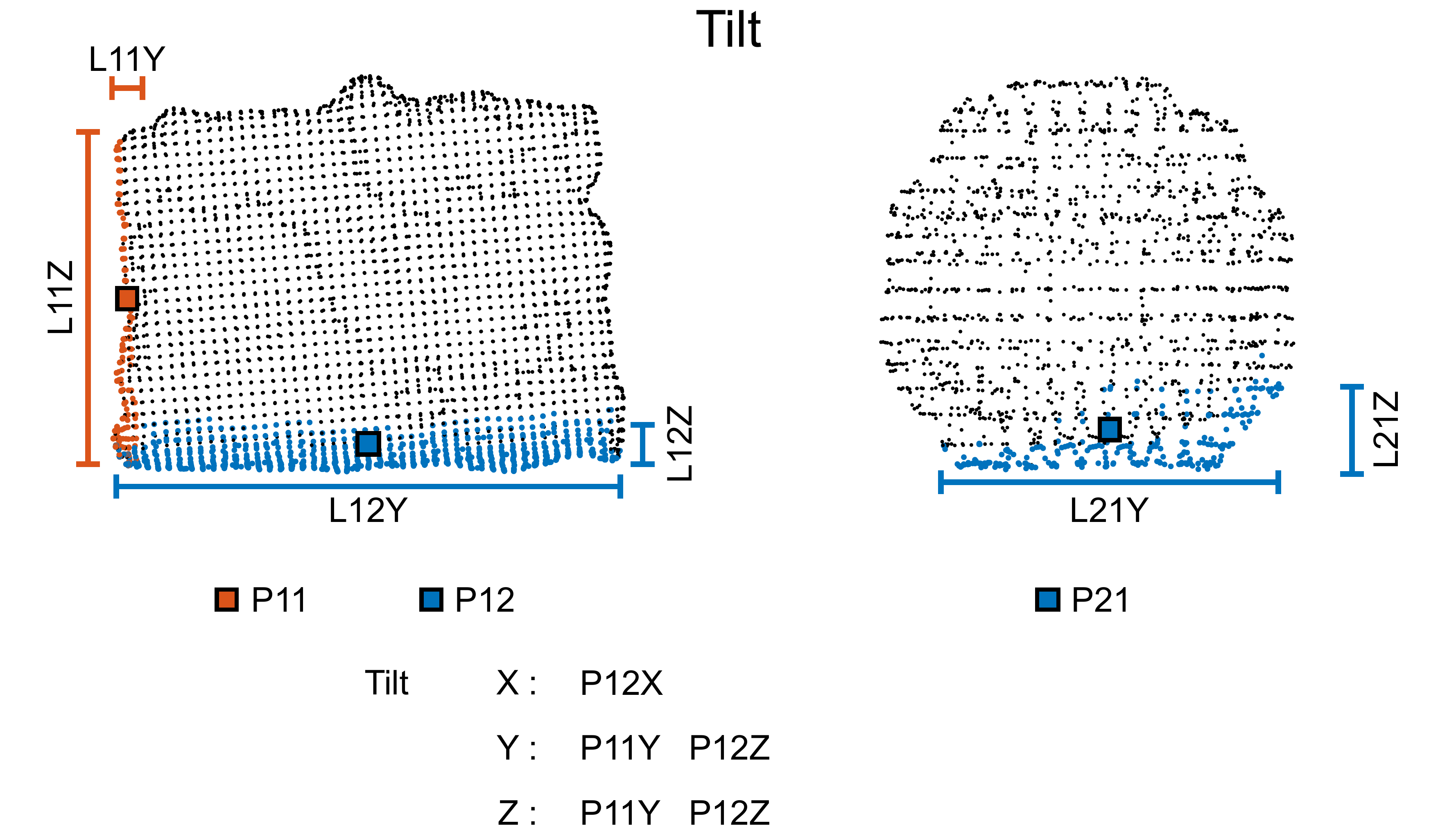}\hspace{0mm}
\includegraphics[width=0.99\columnwidth]{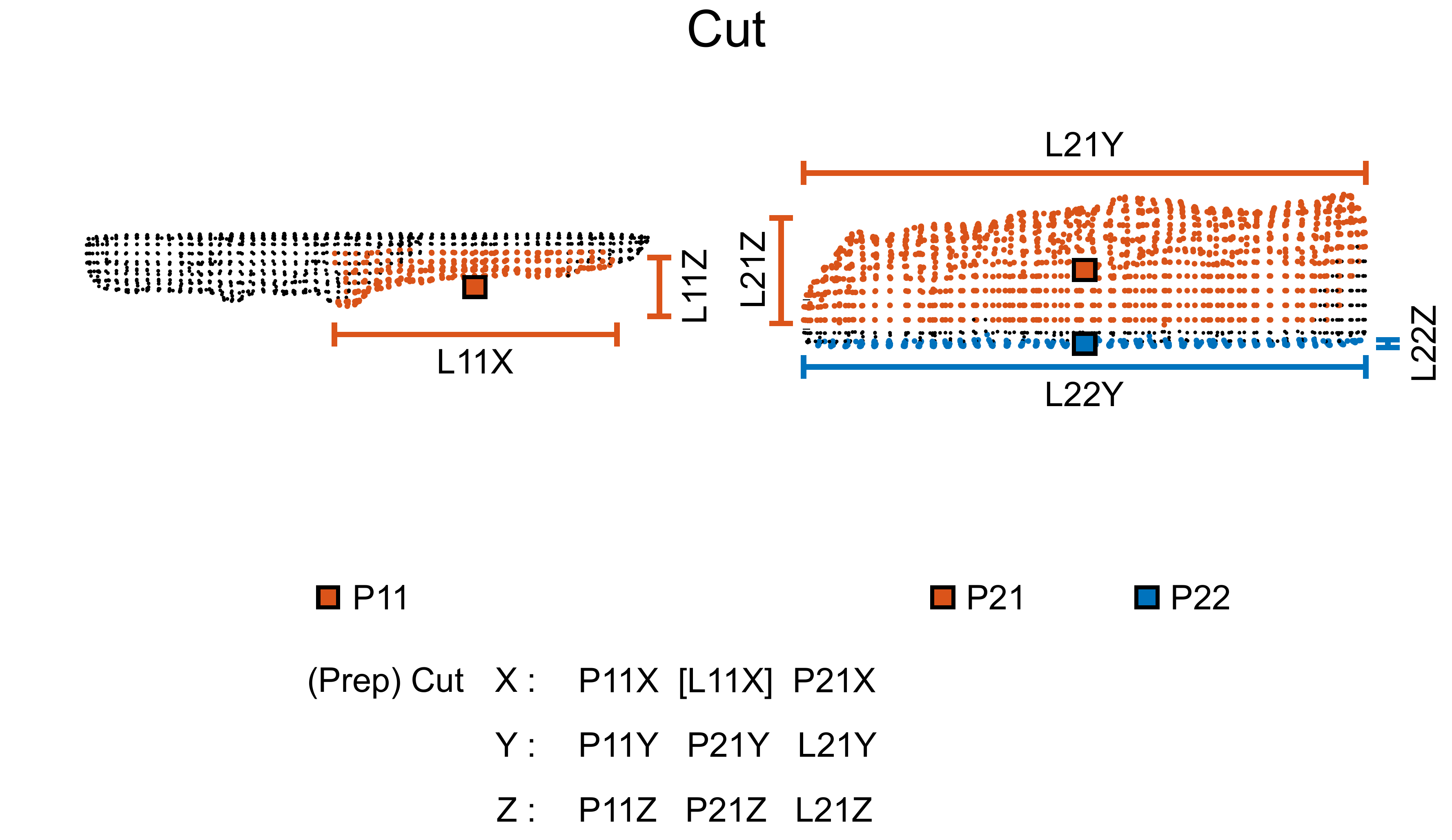}\vspace{10mm}
\includegraphics[width=0.99\columnwidth]{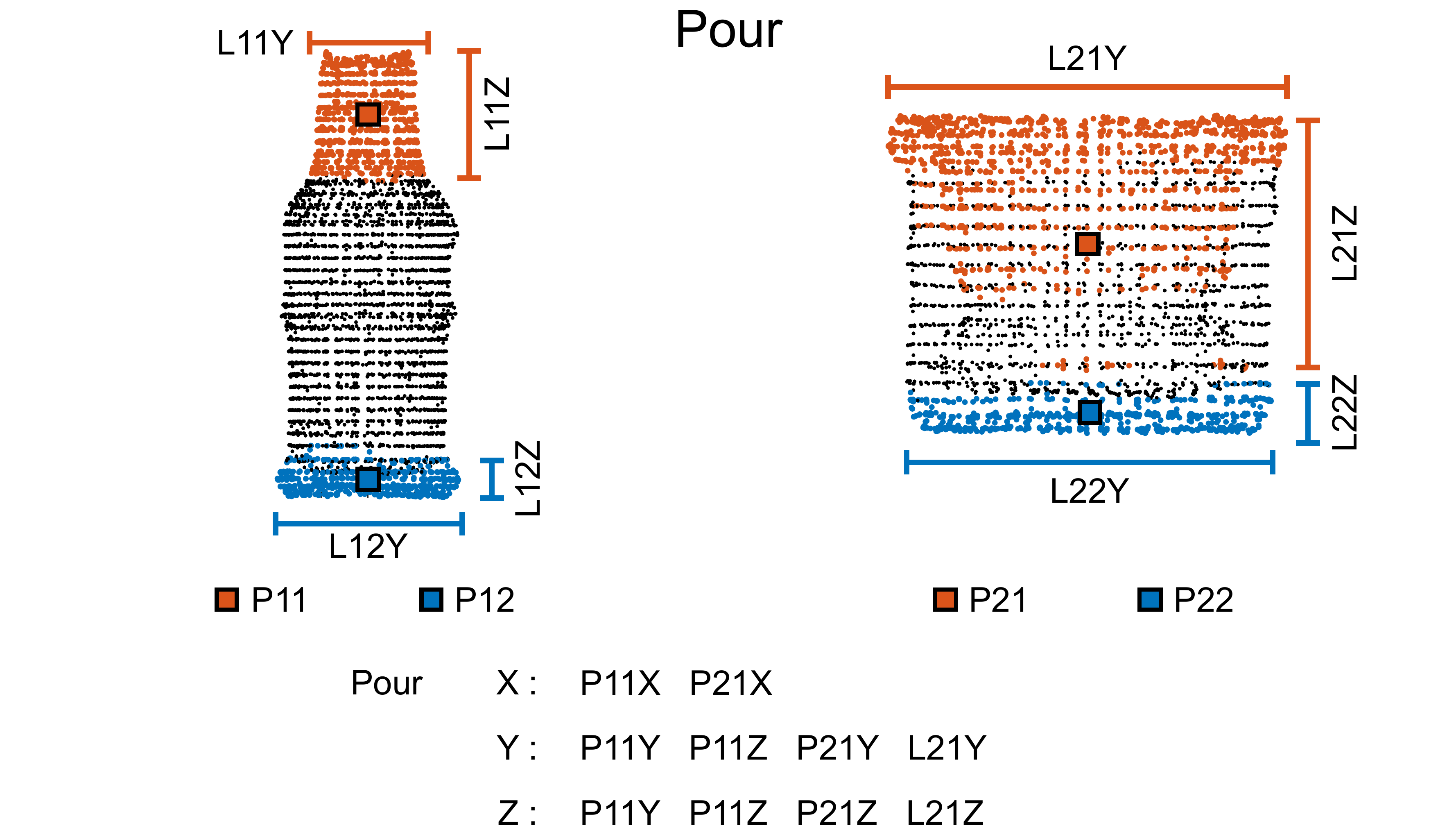}\hspace{0mm}
\includegraphics[width=0.99\columnwidth]{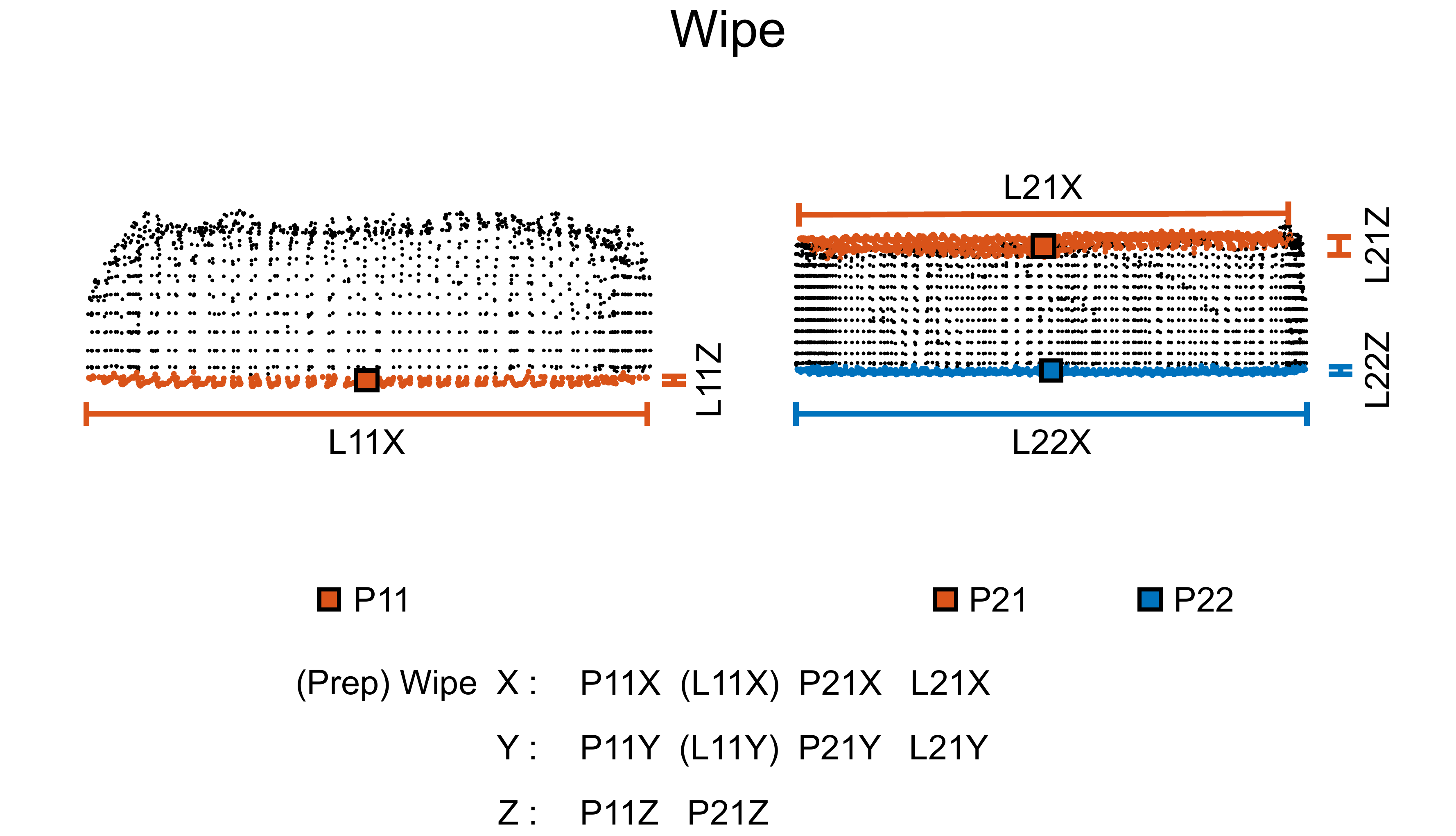}
\par\end{centering}
\protect\caption{\label{fig:selectFeat}	Parts of objects and their corresponding features. The features are aligned with the task frame. The third object in each task is not shown The feature naming convention indicates position/length, object number,  part number, and x-y-z dimension. The human selected features for each skill components are listed below the picture. Features in square brackets only apply to the manipulation skill and features in parenthesis only apply to the preparation skill}
\end{figure*}

Examples of extracted parts for each task are shown in Fig. \ref{fig:object-parts}. 
The figure shows that the demonstration-based part segmentation successfully
detects task relevant parts such as the blade of the knife for cutting,
the opening of the bottle for pouring, and the supporting surface
of the large can for placing. There are some variations in the sizes
and shapes of the detected parts, which will influence the resulting
features. For example, the shaft of the screwdriver, Fig. \ref{fig:object-parts}C,
was incorporated into the tip part for some of the demonstrations.
In order to estimate these variations, we computed the object-relative
positions of the parts and the size of their bounding boxes. We then
computed the standard deviations in these values across the five demonstrations
for each object. The average standard deviation of the values was
$3$mm. It should be noted that the actual features, which are computed
in the task frame and not the object frame, exhibit much greater variations
across the demonstrations.

\subsection{Benchmarking Priors for Feature Selection\label{sub:Benchmarking-Priors-for}}

Given the extracted parts, the next step is to select the relevant
object features for learning the versatile manipulation skill. The goal
of this experiment is to evaluate using a meta-level prior
to transfer knowledge about feature relevance between different skills.
This benchmark experiment therefore compares the performance of the
SSVS when using a uniform prior versus a meta prior. 

Rather than learning one motor primitive per task, the pushing, wiping,
and cutting tasks were each split into two skills. The first skill
corresponds to moving the held object into contact with the second
object. The second skill then performs the pushing, wiping, or cutting
manipulation. We indicate the first skill by \emph{prep} (e.g., cut
prep) as it is used to prepare the objects for the manipulation \cite{Kroemer_ICRA_15}. The
set of relevant object features and the meta features' values
may vary between the skills and the corresponding prep skills. 

The relevant features were selected using the SSVS approach described
in Section \ref{sec:Learning-Relevant-Features}. After the temporal scaling
 of the target DMPs, the rotational components
of the skills exhibited little variation between the different demonstrations,
even for the pouring and tilting tasks. These components were therefore
modeled using a single constant feature, and our feature selection
evaluation was applied to only the three translational components
of the skill. The hyperparameters were set to $a=5$ and $b=5$ for
the prior on the standard deviations $\sigma_{k}$. For the shape
parameters $w_{jk}$, we set the hyperparameters to $\hat{s}^{2}=0.15^{2}$
and $\check{s}^{2}=50\hat{s}^{2}$, such that the prior variances
on the relevant features are fifty times greater than those of the
irrelevant features. The DMPs were learned using five shape parameters
per feature per skill component, resulting in an average of $5\times27\times3=405$
parameters to be learned per skill. As part of the preprocessing,
both the target shape parameters $\tilde{w}_{ik}$ and the object
features $\phi_{ji}$ were normalized across the $N$ training samples. 

\begin{figure}
\begin{centering}
\noun{Feature Selection Accuracy\vspace{0.1mm}
}
\par\end{centering}

\begin{centering}
\includegraphics[width=1\columnwidth]{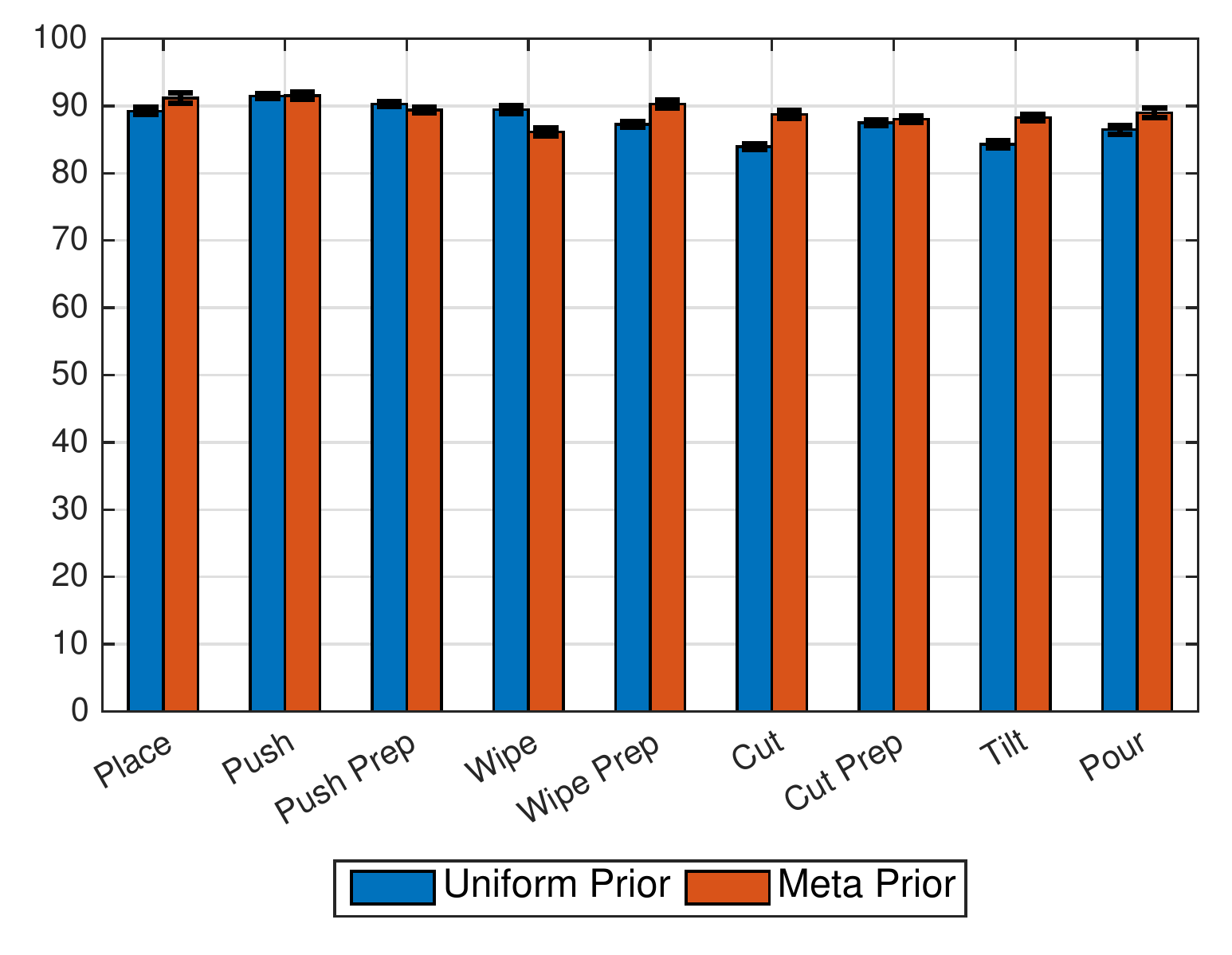}\vspace{-3mm}

\par\end{centering}

\protect\caption{\label{fig:accuracy}The figure shows the accuracy of the SSVS feature
selection process. The blue bar indicates the accuracy when using
a standard uniform prior over the features' relevance. The red bar
shows the the accuracy when using a meta prior that was trained on
the feature relevance of the other tasks' skills. The error bars indicate +/- one
standard deviation. }
\end{figure}

The meta prior's parameters $\theta_{h}$ were trained
using iterative reweighted least squares. The training data was obtained
from the other tasks' skills, i.e., all skills excluding the current skill and the 
corresponding prep skill when applicable. The relevant features of the prior skills
were hand-labelled for the purpose of this evaluation. 
The manually selected features for each skill are shown in Fig. \ref{fig:selectFeat}. 
In ambiguous cases, we selected the features that minimized the leave one out cross validation error for the trajectory prediction. 

The meta-prior training set was created by randomly selecting half of the positive samples $\gamma=1$
and an equal number of negative samples $\gamma=0$ from the pool of previous skills' features. 
The feature relevance prior, as described in
Section \ref{sub:Probablisic-Model-for}, can be computed from a single
demonstration. In order to compute the prior for the entire skill,
we took the average over the priors computed from the meta features
of the individual training demonstrations. 
The uniform prior assumes that each feature has the same a priori probability of being selected, i.e., $p(\gamma_j)=c\forall j\in\{1,...,M\}$ where $c$ is a constant. 
The constant probability was obtained by computing the average number of relevant features 
using all of the features from the previous skills. Over all of the skills, $9.97\%$ of the features are considered to be relevant.

\begin{figure}
\begin{centering}
\noun{Feature Selection Recall\vspace{0.1mm}
}
\par\end{centering}

\begin{centering}
\includegraphics[width=1\columnwidth]{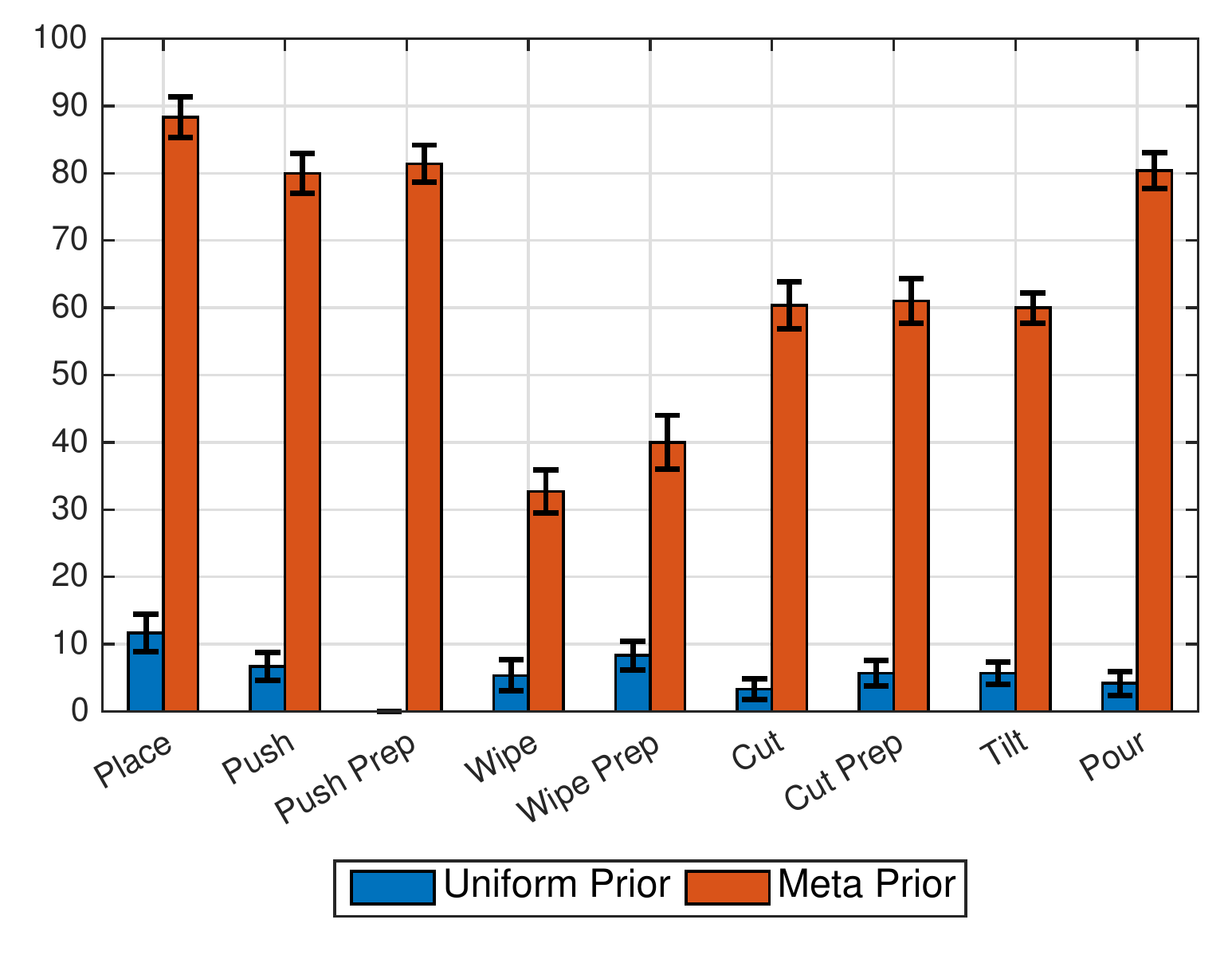}\vspace{-3mm}

\par\end{centering}

\protect\caption{\label{fig:recall}The figure shows the recall of the SSVS feature
selection process. The blue bar indicates the recall when using a
standard uniform prior over the features' relevance. The red bar shows
the the recall when using a meta prior that was trained on the feature relevance of the other tasks' skills. 
The error bars indicate +/- one standard
deviation. }
\end{figure}
\begin{figure}
\begin{centering}
\noun{Feature Selection Precision\vspace{0.1mm}
}\includegraphics[width=1\columnwidth]{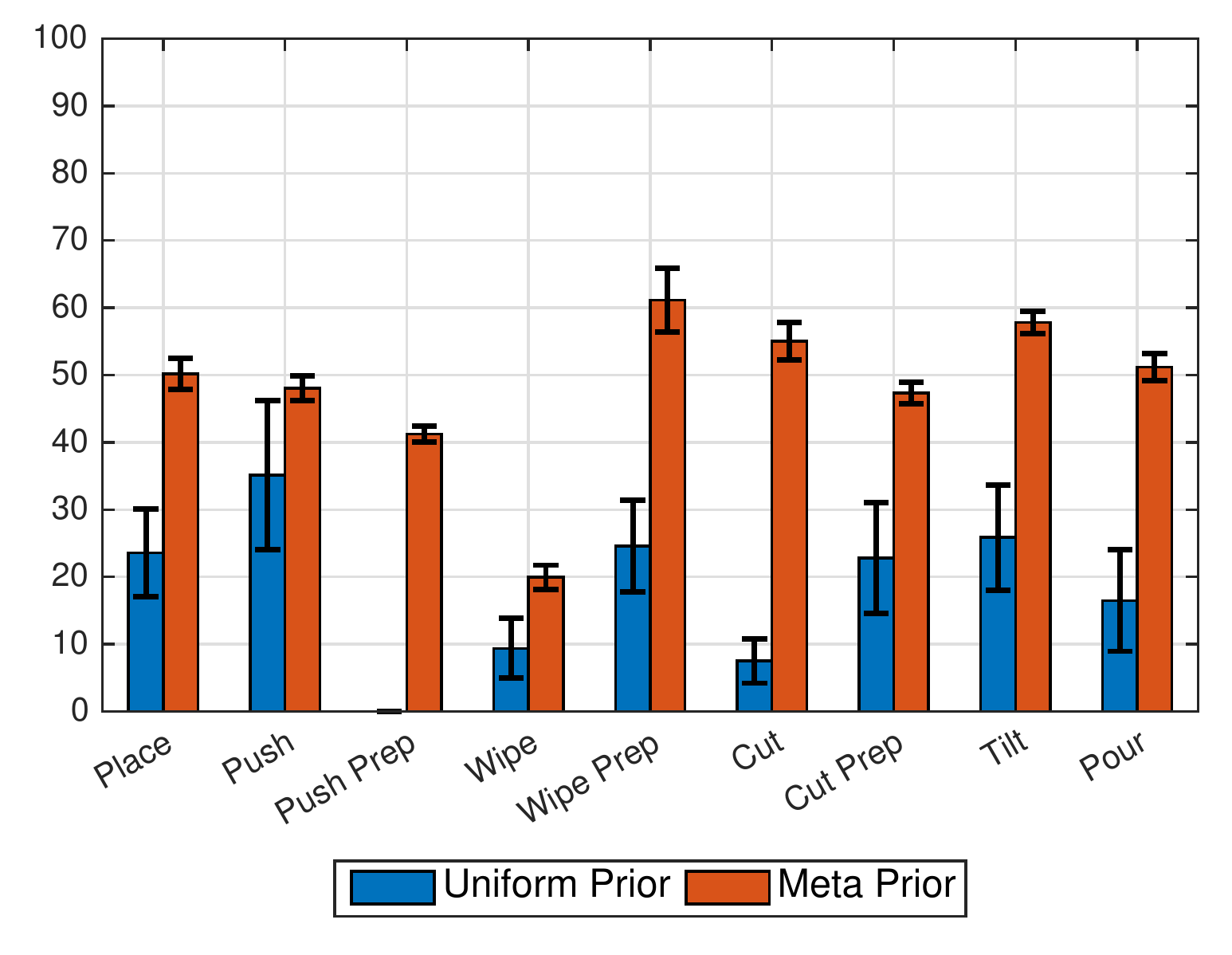}\vspace{-3mm}

\par\end{centering}

\protect\caption{\label{fig:precision}The figure shows the precision of the SSVS feature
selection process. The blue bar indicates the precision when using
a standard uniform prior over the features' relevance. The red bar
shows the the precision when using a meta prior that was trained on
the feature relevance of the other tasks' skills. The error bars indicate +/- one
standard deviation. }
\end{figure}

\begin{figure*}
\begin{centering}
\noun{Goal Prediction Error (cm)\vspace{0.1mm}}
\par\end{centering}
\begin{centering}
\includegraphics[width=1\columnwidth]{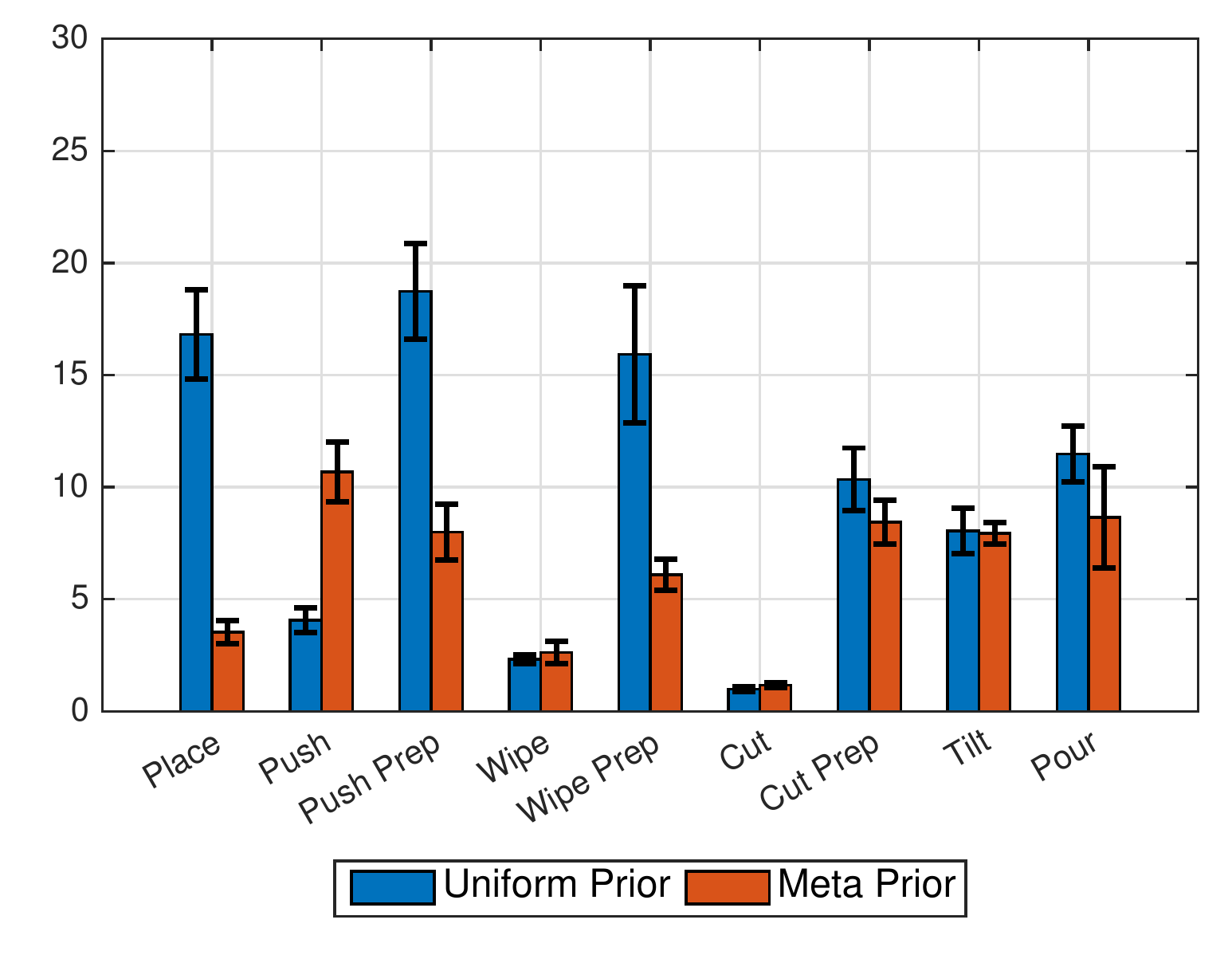}
\includegraphics[width=1\columnwidth]{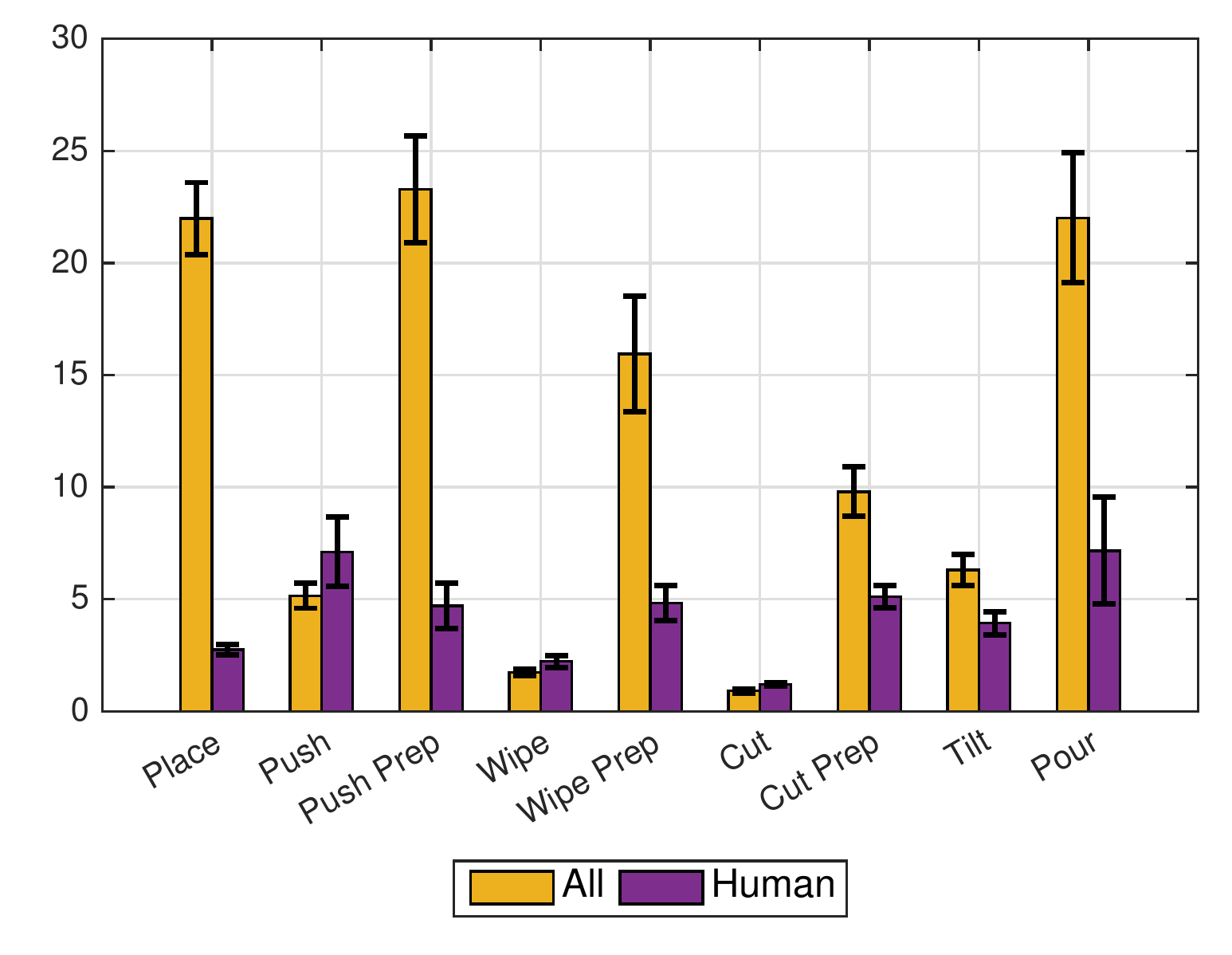}\vspace{-3mm}
\par\end{centering}
\protect\caption{\label{fig:GoalPred} The plots show the average errors in centimeters for the goal prediction evaluation. The left plot shows the results when using SSVS with a uniform prior and a meta-level prior respectively. The plot on the right shows the two baseline methods of using all of the features and human selected features. The errorbars indicate +/- one standard error.}
\end{figure*}

The set of relevant features were estimated using the Gibbs sampling
approach outlined is Section \ref{sub:Inferring-Relevant-Object}.
For each trial, the robot randomly selected $N=5$ demonstrations from the current
skill. The SSVS approach was applied to these five trials.
All of the relevance parameters were initialized to one $\gamma_{j}=1$.
We used $200$ samples for the burn-in stage of sampling, followed
by another $1000$ samples for the actual sampling. Using a Bayes
estimator approach, a feature was considered relevant iff it was assigned
a value of $\gamma_{j}=1$ for the majority of the samples. These
estimates of the relevant features were then compared to the ground
truth human labels. 
The results for using the meta prior, as well
as the uniform prior are shown in Fig. \ref{fig:accuracy}, \ref{fig:recall},
and \ref{fig:precision}. The evaluations were repeated ten times
for each skill, with different training sets for the meta priors and demonstrations for the feature selection.
The errorbars indicate the standard deviations over these evaluations.

Both of the priors result in high accuracies for the feature selection,
with the meta prior approach having an average accuracy of $89.2\%$.
However, due to the sparsity of the relevant features, a high accuracy
can be achieved by treating all of the features as irrelevant $\gamma_{j}=0\forall j\in\{1,...M\}$.
The resulting manipulation skills would however not generalize between
different situations. We therefore need to consider the precision
and recall to determine how well the two models capture the relevant
features.

The recall is shown in Fig. \ref{fig:recall}. While the meta prior
approach achieves an average recall of $64.9\%$, the standard uniform
prior only has an average recall of $5.7\%$.  The recall values can be
improved by including additional demonstrations. This result indicates
that the meta-level prior guides the sampling process to capture most
of the relevant features. In contrast, given only a few demonstrations, 
many of the relevant features are missed when using a uniform prior.
A high recall is particularly important if the robot subsequently uses reinforcement learning to
optimize the skill, as the robot is less likely to explore variations
in features that it already considers irrelevant. 

The benefit of the meta prior can also be seen in the precision, as
it increase the average precision from $18.36\%$ to $48.9\%$. The 
precision values are lower due to the lack of variations in
the demonstrations. For example, the vertical position of the bottom
of the grasped object is a relevant feature for the place task. However,
as all of the objects are initially resting on the table, the bottoms
of all three objects are at the same height. As a result, the Gibbs
sampling often considered these features to be relevant as well, which
allowed it to use them to estimate the height of the table more accurately.
However, if the task were performed with the objects placed on different
surfaces, then these additional feature would lead to worse generalization.
The precision values could therefore be improved through more samples,
either from additional demonstrations or through self exploration,
in novel scenarios. Once the robot has mastered a skill, the set of selected features could
be used to update the meta prior for future tasks. 

As a benchmarking experiment, the results show that the meta-level prior more than
doubles the precision  and recall of the feature selection compared to the standard
uniform prior. The results thus show that the meta prior can autonomously 
transfer prior relevance knowledge between distinct skills.

\subsection{Goal State Prediction \label{sub:GoalPred}}

\begin{figure*}
\begin{centering}
\includegraphics[width=0.166\linewidth]{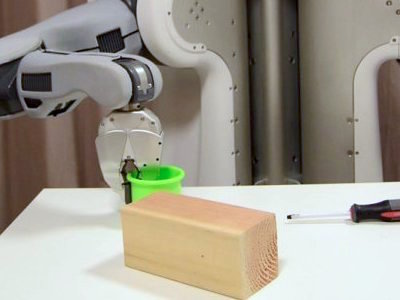}\includegraphics[width=0.166\linewidth]{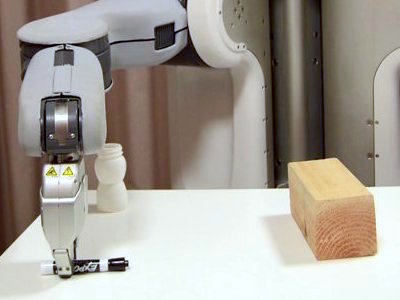}\includegraphics[width=0.166\linewidth]{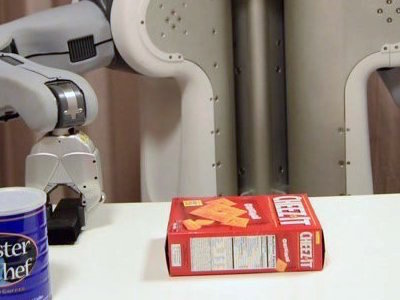}\includegraphics[width=0.166\linewidth]{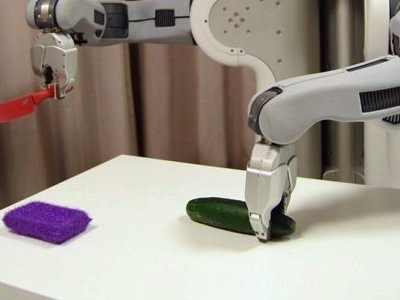}\includegraphics[width=0.166\linewidth]{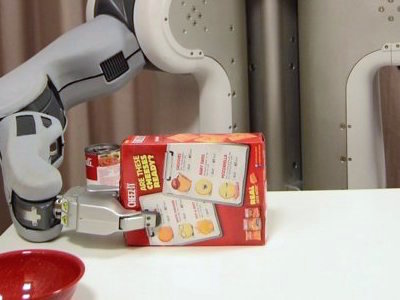}\includegraphics[width=0.166\linewidth]{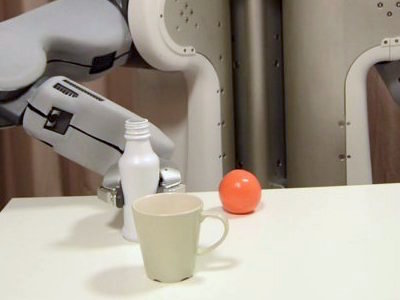}
\par\end{centering}

\centering{}\includegraphics[width=0.166\linewidth]{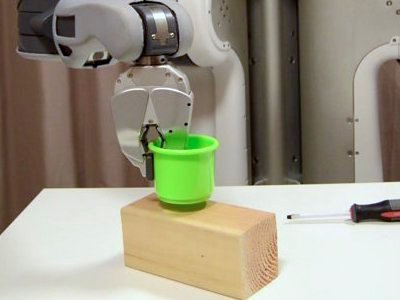}\includegraphics[width=0.166\linewidth]{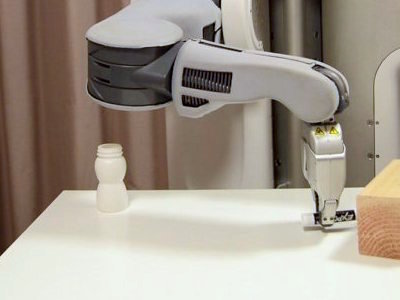}\includegraphics[width=0.166\linewidth]{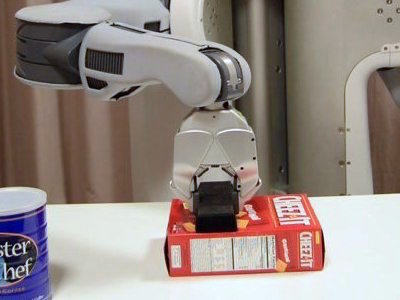}\includegraphics[width=0.166\linewidth]{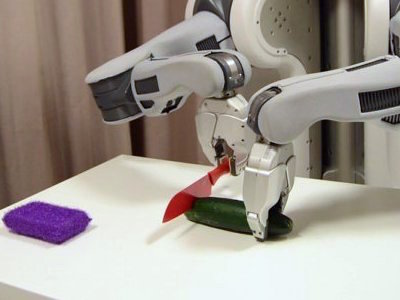}\includegraphics[width=0.166\linewidth]{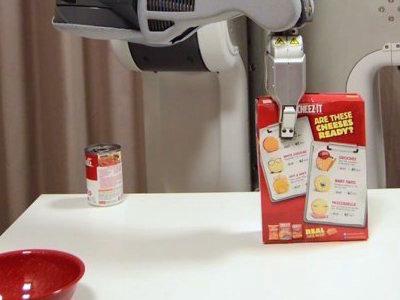}\includegraphics[width=0.166\linewidth]{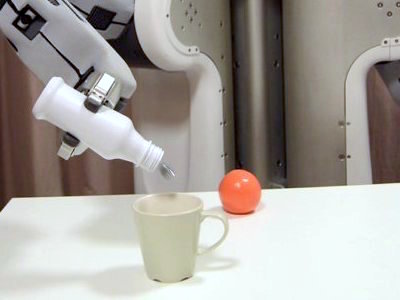}\protect\caption{\label{fig:Pictures-of-Execution}The pictures show the robot's skill executions. The top row shows the initial states for the task. The bottom row
shows the resulting manipulations being performed. From left to right,
the columns show executions from the placing, pushing, wiping, cutting,
tilting, and pouring tasks}
\end{figure*}

In our reformulation of the DMPs, the goal state of the movement is defined as a linear combination of the extracted object features. The robot is therefore also selecting features for predicting the goal state of the movement. 
In this experiment, we evaluated the robot's ability to predict the goal state of the skill from the object features. We again compare using a uniform prior to the proposed meta-level prior. We additionally compare to using all of the features, as well as using the manually selected features. 

In each trial, one of the demonstrations was selected as the test sample, and $N=5$ other demonstrations from the skill were randomly selected as the training samples. For the uniform prior and meta prior approaches, the robot  selected a relevant set of object features for generalizing the skills using SSVS, as described in Section \ref{sec:Learning-Relevant-Features}. The robot then used the selected features to predict the goal state of the test DMP. The root mean squared error (RMSE) between the predicted and the demonstrated goal states were computed to evaluate the approaches.  

The results of the experiment are shown in Fig. \ref{fig:GoalPred}. We have included the results for all of the tasks for completeness, although not all of the tasks require a specific goal location. For example, the pushing distance varied randomly between demonstrations. The average errors over all of the skills are $4.33$cm for human-selected features, $6.34$cm for meta-prior features, $9.86$cm for uniform-prior features, and $11.9$cm for all of the features.

The meta and uniform priors achieved similar results for the wiping, cutting, and tilting tasks. These tasks require relatively small movements and the skills therefore do not need to adapt to large variations in the object features.
The meta prior outperformed the uniform prior for the placing, pouring, and prep skills. These skills involve moving the grasped object into close proximity to a part of another object. 
In addition to requiring larger movements, the object features for these tasks also exhibit more variations as the initial scenes are less constrained than in the cutting, wiping, and tilting tasks. As a result, selecting the wrong features can lead to larger errors. The meta-level prior guides the robot in selecting the relevant features given the limited number of training demonstrations. 

The benefit of the meta-level prior is also greater for these tasks due to the similarities in the skills' relevant features, i.e., the relevant features for these skills are generally axis-aligned and in close proximity to the hand at the end of the demonstration. The meta-level prior is therefore better at predicting the relevant features as it has more training samples. In contrast, the tilting task includes relevant features across dimensions, e.g., the y component of the skill depends on the z position of a part. Pouring is the only other skill from the set that includes these kinds of relevant features. As a result, the benefit of the meta-level prior may be smaller for these skills, but it could be increased by learning additional skills. 

Pushing is the only skill for which the meta-level prior performed substantially worse than the uniform prior. The manually selected features also performed worse for this task. This result demonstrates that an erroneous prior can also decrease performance. The meta-level prior and the human both expected certain features to be relevant, and the robot therefore attempted to model the random variations in the pushing distance based on these features. 
The influence of the prior will decrease as the robot obtains more training samples.  

As one would expect given the limited amount of training data, the manually selected features have the best performance. The meta-prior approaches performance could be further improved by using additional meta-features to differentiate between relevant and irrelevant features, as well as learning from more previous skills. Overall, the experiments have shown that the meta-level prior, in comparison to the uniform prior or no feature selection, can substantially increase the performance of skills learned from only a few demonstrations.

\subsection{Learning Manipulation Skills \label{sub:Learning-Manipulation-Skills}}

In the final experiment, the learned skills were executed on the real
robot to verify that the skills can adapt to different situations
and perform the intended manipulations. The skills were trained using
the relevant set of features learned using
the meta prior. The skills were learned using all $N=15$ demonstrations.
In order to test the skills with different object features,
the evaluations were performed using new combinations of objects,
in varying positions and orientations, and with different grasps.
Each skill was executed three times. If a prep skill failed
in such a manner that the subsequent skill could not be performed,
then it was marked as a failure and another execution was performed
to evaluate the following manipulation skill. 

Examples of the skill executions can be seen in Fig. \ref{fig:Pictures-of-Execution},
as well as in the attached video. The video also includes examples
of failed skill executions. The robot achieved an overall success
rate of $77.8\%$ during the evaluations. Learning these skills from
$N=15$ demonstrations is not a trivial task, as each skill had $375$
to $465$ parameters to be learned. Therefore, although the skills
capture the coarse movements that allow the robot to perform the manipulations,
the executions could be further refined and made more robust. In
particular, the amplitudes of the pushing and cutting movements tended
to be slightly too large, the pouring was performed off-center, the tilting resulted in
horizontal shifts of the boxes, and half of the failed executions were the result of prep skills
stopping a couple of centimeters before making contact. These issues could be alleviated
by adjusting the robot's trajectories by a few centimeters. 

This skill refinement could be achieved through additional self exploration
using reinforcement learning. As the skills only require minor adjustments,
the robot could employ a policy search approach \cite{Pastor_ICRA_2011,Levine_ICRA_2015}. The skill executions could also be refined by adding
visual/tactile/haptic feedback into the DMPs \cite{Pastor_Humanoids_2012,Chebotar_IROS_2014,Yamaguchi_Humanoids_2015}.
These feedback terms can be easily incorporated as features $\phi$
that depend on the current state rather than the initial state. The
feature selection process could thus also be extended to select relevant
feedback terms.

The experiments have shown that the robot can use the SSVS approach
together with a meta-level prior to efficiently learn versatile
manipulations from relatively few demonstrations. Having demonstrated
the feasibility and benefits of this approach, we plan on extending
the skill learning framework in the future to allow for different
action frames and larger variations in object orientations. We will
also investigate explicitly incorporating the interaction sites in
the feature generation process. These interactions, which are currently
used to initialize the part segmentation, could provide additional
details for generalizing manipulation skills.

\section{Conclusion\label{sec:Conclusion}}

In this paper, we presented a learning-from-demonstration method for
selecting the relevant object features for generalizing manipulation
skills. The proposed approach is based on stochastic search variable
selection (SSVS). We extend the standard SSVS model
to incorporate a meta-level prior, which allows the robot to transfer
knowledge regarding feature relevance between different skills. The transfer was
performed using meta features that capture the proximity and alignment
of the object features. 
We also explained how the GrabCut
segmentation method can be used to extract affordance-bearing parts
of objects from demonstrations. The extracted parts formed the
basis for the automatic feature generation.

The proposed method was evaluated using a PR2 robot. The robot learned placing,
pushing, cutting, wiping, tilting, and pouring skills that generalize
between different sets of objects and object configurations. The results
show that the meta prior allows the robot to more than double the
feature selection's average precision and recall compared to
a uniform prior.
 \bibliographystyle{plainnat}
\bibliography{KroemerMeta2016}

\end{document}